\documentclass[acmsmall]{acmart}

\usepackage{xurl}
\usepackage{makecell}
\renewcommand{\arraystretch}{1.3}
\usepackage{algorithmic}
\usepackage{graphicx}
\usepackage{textcomp}
\usepackage{xcolor}
\usepackage{multirow}
\usepackage{tikz}
\usetikzlibrary{positioning}
\usetikzlibrary{math}
\usepackage{pgfplots}

\def\BibTeX{{\rm B\kern-.05em{\sc i\kern-.025em b}\kern-.08em
    T\kern-.1667em\lower.7ex\hbox{E}\kern-.125emX}}
\usepackage{pgfplots}
\pgfplotsset{compat=1.7}

\begin{document}

\title{A Survey of Deep Learning and Foundation Models for Time Series Forecasting}

\author{John A. Miller}
\email{jamill@uga.edu}

\author{Mohammed Aldosari}
\email{maa25321@uga.edu}

\author{Farah Saeed}
\email{farah.saeed@uga.edu}

\author{Nasid Habib Barna}
\email{nasidhabib.barna@uga.edu}

\author{Subas Rana}
\email{subas.rana@uga.edu}

\author{I. Budak Arpinar}
\email{budak@uga.edu}

\author{Ninghao Liu}
\email{ninghao.liu@uga.edu}

\begin{abstract}
Deep Learning has been successfully applied to many application domains, yet its advantages have been slow to emerge for time series forecasting.
For example, in the well-known Makridakis (M) Competitions, hybrids of traditional statistical or machine learning techniques have only recently become the top performers.
With the recent architectural advances in deep learning being applied to time series forecasting
(e.g., encoder-decoders with attention, transformers, and graph neural networks), deep learning has begun to show significant advantages.
Still, in the area of pandemic prediction, there remain challenges for deep learning models: the time series is not long enough for effective training, unawareness of accumulated scientific knowledge, and interpretability of the model.
To this end, the development of foundation models (large deep learning models with extensive pre-training) allows models to understand patterns and acquire knowledge that can be applied to new related problems before extensive training data becomes available.
Furthermore, there is a vast amount of knowledge available that deep learning models can tap into, including Knowledge Graphs and Large Language Models fine-tuned with scientific domain knowledge.
There is ongoing research examining how to utilize or inject such knowledge into deep learning models.
In this survey, several state-of-the-art modeling techniques are reviewed, and suggestions for further work are provided.
\end{abstract}

\maketitle


\section{Introduction}

The experience with COVID-19 over the past four years
has made it clear to organizations such as the National Science Foundation (NSF)
and the Centers for Disease and Prevention (CDC) that
we need to be better prepared for the next pandemic.
COVID-19 has had huge impacts with 6,727,163 hospitalizations and 1,169,666 deaths
as of Saturday, January 13, 2024, in the United States alone
(first US case 1/15/2020, first US death 2/29/2020).
The next one could be more virulent with greater impacts.

There were some remarkable successes such as the use of messenger RNA vaccines
that could be developed much more rapidly than prior approaches.
However, the track record for detecting the start of a pandemic
and the forecasting of its trajectory leaves room for improvement.

Pandemic Preparedness encapsulates the need for continuous monitoring. 
Predicting rare events in complex, stochastic systems is very difficult.
Transitions from pre-emergence to epidemic to pandemic are easy to see
only after the fact.
Pandemic prediction using models is critically important as well.
Sophisticated models are used to predict the future of hurricanes
due to their high impact and potential for loss of life.
The impacts of pandemics are likely to be far greater.

As with weather forecasting, accurate pandemic forecasting
requires three things:
(1) a collection of models,
(2) accurate data collection, and
(3) data assimilation.
If any of these three break down, accuracy drops.
When accuracy drops, interventions and control mechanisms
cannot be optimally applied, leading to frustration from the public.

\begin{figure}[htbp]
    \includegraphics[width=0.7\textwidth, height=1.9in]{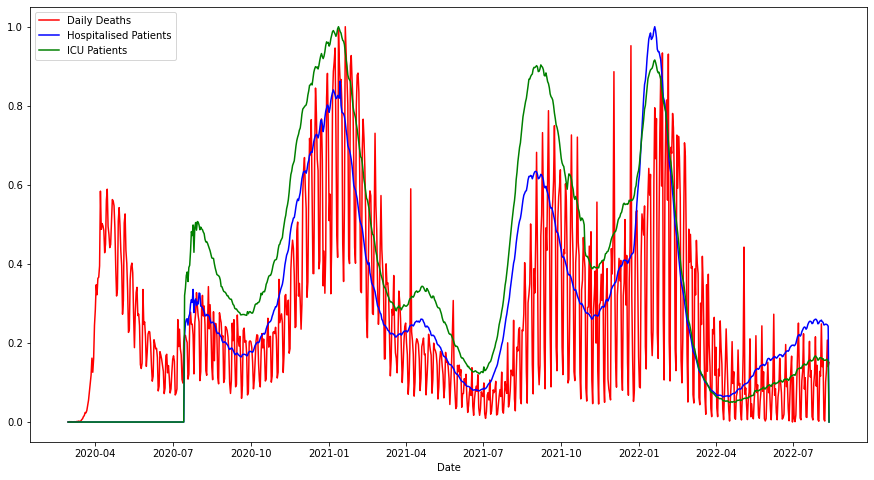}
    \caption{Rescaled Plot of Daily Deaths during the Active Pandemic (note the weekly reporting pattern), Hospitalization, and ICU Patients.}
    \label{fig:dailydeath}
\end{figure}

During the COVID-19 pandemic data were collected daily, but as seen in Figure \ref{fig:dailydeath},
there is a very strong weekly pattern that dominates the curve
of new deaths that is an artifact of reporting processes.
Also, notice how strong hospitalizations and the number of Intensive Care Unit (ICU) patients
appear to be good leading indicators.

Due to the saw-tooth pattern in the daily deaths, some modeling studies find it better to work off of weekly data.
In the later stages of COVID-19, daily reporting stopped, and only weekly remains.
Unfortunately, this means that there is much less data available for training deep learning models.

Modeling techniques applied were statistical, machine learning or
theory-based compartmental models that were extensions to Susceptible-Infected-Recovered (SIR) or
Susceptible-Exposed-Infected-Recovered (SEIR) models.
Transitions among these states are governed by differential equations
with rate constants that can be estimated from data.
Unfortunately, estimating the population of individuals that are in, for example,
the exposed state can be very difficult.

The other two categories, statistical and machine learning (including deep learning and foundation models),
one could argue are more adaptable to available data as they look for patterns that repeat,
dependency on the past, and leading indicators.
Both can be formulated as Multivariate Time Series (MTS) forecasting problems,
although the related problems of MTS classification and anomaly detection
are also very important.
Still, having a connection to theory is desirable and could lead to better forecasting
in the longer term, as well as a greater understanding of the phenomena.
This has led to work in Theory-Guided Data Science (TGDS)
\cite{karpatne2017theory, miller2017adding} and
Physics-Informed Neural Networks (PINN) \cite{karniadakis2021physics}.

Statistical and machine learning techniques complement each other.
For example, modeling studies should have reliable baseline models
that, from our studies, should include Random Walk (RW), Auto-Regressive (AR), and
Seasonal, Auto-Regressive, Integrated, Moving Average with eXogenous variables (SARIMAX).
SARIMAX is typically competitive with Deep Learning models
when training data are limited.
If weekly data are used, the training data will be limited for much of the early
stages of the pandemic, just when the need for accurate forecasts is the greatest.
Baselines like SARIMAX can also be helpful for hyper-parameter tuning,
in that with sufficient data, one would expect deep learning models to perform well;
the SARIMAX results can help gauge this.
Furthermore, SARIMAX has been used for data augmentation to help train deep learning
models \cite{javeri2021improving}.

Looking toward the future, this survey paper that extends \cite{miller2023knowledge}
asks the question of how Artificial Intelligence (AI), particularly deep learning,
can be used to improve pandemic preparedness and prediction,
in terms of better deep learning models, more explainable models,
access to scientific literature using Large Language Models (LLM),
development and use of knowledge bases and knowledge graphs,
as well as better and ongoing assessment of pandemic interventions and controls.

The rest of this paper is organized as follows:
Section 2 provides an overview of two waves of improvement in MTS forecasting.
Section 3 focuses on recent progress in MTS forecasting looking at Transformers
and related modeling techniques.
These modeling techniques increasingly strive to better capture temporal dynamics
and tend to be the top performers for national-level COVID-19 forecasting.
Section 4 focuses on recent progress in MTS forecasting in the spatial-temporal 
domain, where various types of Graph Neural Networks have a natural appeal.
These modeling techniques tend to be applied to state-level COVID-19 data.
Foundation models, large pre-trained deep learning models,
for time series forecasting are discussed in Section 5.
Knowledge in various forms such as Knowledge Graphs is a natural complement
to forecasting models, as discussed in Section 6.
The knowledge can be used to improve forecasting accuracy, check the reasonableness
for forecasts (especially a problem for long-term forecasting), guide the
modeling process, and help explain the modeling results.
A meta-study comparing the effectiveness of several modeling techniques found in the current literature
is given in section 7.
Finally, a summary is given in Section 8 that includes looking into a crystal ball
to see where MTS might be heading in the future.

\section{Progress in MTS Forecasting}

The history of time series forecasting dates way back as shown in Table \ref{tab:ts_papers}.
Note that some of the modeling techniques (or model types) have general use other than time series
forecasting, but the date and paper reflect their use in forecasting for time series or sequence data.
There were, however, periods of rapid progress, for example, the one in the 1950s through the '70s,
and captured in the seminal book by Box and Jenkins \cite{box1970time}.
Another period of substantial progress coincides with the advancement of deep learning,
in the 2010s.

\begin{table}[htbp]
\caption{Types of Time-Series Forecasting Models with Key Initial Reference in Time-Series Context}
\setlength{\tabcolsep}{0.5pt}
\begin{center}
\begin{tabular}{|c|c|c|c|} \hline
{\bf Model Type} & {\bf Short Description} & {\bf Date} & {\bf Reference} \\ \hline \hline
BaselineHA     & Historical Average Baseline & . & . \\ \hline
BaselineRW     & \makecell{Random Walk Baseline -\\ guess the previous value} & . & . \\ \hline
Regression4TS  & \makecell{Time-Series Regression\\with Lagged Variables} & 1949 & \cite{cochrane1949application} \\ \hline
ARIMA          & AutoRegressive, Integrated, Moving-Average & 1951 & \cite{whittle1951hypothesis, box1962some} \\ \hline
SES            & Simple Exponential Smoothing & 1957 & \cite{holt1957forecasting} \\ \hline
VAR            & Vector AutoRegressive & 1957 & \cite{quenouille1957analysis} \\ \hline
VARMA          & Vector AutoRegressive, Moving-Average & 1957 & \cite{quenouille1957analysis} \\ \hline
VECM           & Vector Error Correction Model & 1957 & \cite{quenouille1957analysis} \\ \hline
ES             & Exponential Smoothing (Holt-Winters) & 1960 & \cite{winters1960forecasting} \\ \hline
SARIMA         & Seasonal ARIMA & 1967 & \cite{box1967models} \\ \hline
SARIMAX        & SARIMA, with eXogenous variables & 1970 & \cite{box1970time} \\ \hline
NAR/NARX       & Nonlinear AutoRegressive, Exogenous & 1978 & \cite{jones1978nonlinear} \\ \hline \hline

Neural Network & Neural Network & 1988 & \cite{smith1988using} \\ \hline
RNN            & Recurrent Neural Network & 1989 & \cite{williams1989learning} \\ \hline
FDA/FTSA       & Functional Data/Time-Series Analysis & 1991 & \cite{ramsay1991some} \\ \hline
CNN            & Convolutional Neural Network & 1995 & \cite{lecun1995convolutional} \\ \hline
SVR            & Support Vector Regression & 1997 & \cite{muller1997predicting} \\ \hline
LSTM           & Long, Short-Term Memory & 1997 & \cite{hochreiter1997long} \\ \hline
ELM            & Extreme Learning Machine & 2007 & \cite{singh2007application} \\ \hline
GRU            & Gated Recurrent Unit & 2014 & \cite{cho2014learning} \\ \hline
Encoder-Decoder & Encoder-Decoder with Attention & 2014 & \cite{chorowski2014end} \\ \hline
MGU            & Minimal Gated Unit & 2016 & \cite{zhou2016minimal} \\ \hline
TCN            & Temporal Convolutional Network & 2016 & \cite{lea2016temporal} \\ \hline
GNN/GCN        & Graph Neural/Convolutional Network & 2016 & \cite{kipf2016semi} \\ \hline
vTRF           & (Vanilla) Transformer & 2017 & \cite{vaswani2017attention} \\ \hline
\end{tabular}
\label{tab:ts_papers}
\end{center}
\end{table}

There are several open source projects that support time series analysis across programming languages:

\begin{itemize}
\item
{\bf R}:
Time Series [{\url https://cran.r-project.org/web/views/TimeSeries.html}]
consists of a large collection of time series models.
\item
{\bf Python}:
Statsmodels [{\url https://www.statsmodels.org/stable/index.html}]
provides a basic collection of time series models.
Sklearn [\url {https://scikit-learn.org/stable}]
and a related package, Sktime [\url {https://www.sktime.net/en/stable}],
provide most of the models offered by Statsmodels.
PyTorch-Forecasting [{\url https://github.com/jdb78/pytorch-forecasting?tab=readme-ov-file}]
includes several types of Recurrent Neural Networks.
TSlib [\url{https://github.com/thuml/Time-Series-Library}]
provides several types of Transformers used for time series analysis.
\item
{\bf Scala}:
Apache Spark [\url{https://spark.apache.org/}]
has a limited collection of time series models.
{\sc ScalaTion} [\url{https://cobweb.cs.uga.edu/~jam/scalation.html}],
[\url{https://github.com/scalation}] \cite{miller2010using}
supports most of the modeling techniques listed in Table \ref{tab:ts_papers}.
In addition, it supports several forms of Time-Series Regression,
including recursive {\tt ARX}, direct {\tt ARX\_MV}, quadratic, recursive {\tt ARX\_Quad}, and
quadratic, direct {\tt ARX\_Quad\_MV}.
\end{itemize}

Forecasting the future is a very difficult thing to do and until
recently, machine learning models did not offer much beyond what statistical
models provided, as born out in the M Competitions.

\subsection{M Competitions}

The Makridakis or M Competitions began in 1982 and have continued with M6 which took place 2022-2023.
In the M4 competition ending in May 2018, the purely ML techniques performed poorly,
although a hybrid neural network (LSTM) statistical (ES) technique \cite{smyl2020hybrid}
was the winner \cite{makridakis2018m4}.
The rest of the top performers were combinations of statistical techniques.
Not until the M5 competition ended in June 2020, did Machine Learning (ML) modeling techniques
become better than classical statistical techniques.
Several top performers included LightGBM in their combinations \cite{makridakis2022m5}.
LightGBM is a highly efficient implementation of Gradient Boosting Decision/Regression Trees \cite{ke2017lightgbm}.
Multiple teams included neural networks in their combinations as well. 
In particular, deep learning techniques such as
DeepAR consisting of multiple LSTM units \cite{salinas2020deepar} and
N-BEATS, consisting of multiple Fully Connected Neural Network blocks \cite{oreshkin2019n} were applied.
Many of them also combined recursive (e.g., uses $t+1$ prior forecast for $t+2$ forecast) and
direct (non-recursive) forecasting.
``In M4, only two sophisticated methods were found to be more accurate than simple
statistical methods, where the latter occupied the top positions in the competition.
By contrast, all 50 top-performing methods were based on ML in M5". \cite{makridakis2022m5}
The M6 competition involves both forecasting and investment strategies \cite{makridakis2023m6}
with summaries of its results expected in 2024.

\subsection{Statistical and Deep Learning Models for Time Series}

As illustrated by the discussion of the M Competitions, machine learning
techniques took some time to reach the top of the competition.
Neural Network models demonstrated highly competitive results in many
domains, but less so in the domain of time series forecasting,
perhaps because the patterns are more elusive and often changing.
Furthermore, until the big data revolution, the datasets were too small
to train a neural network having a large number of parameters.

\subsubsection{First Wave}

From the first wave of progress, SARIMAX models have been shown
to generally perform well, as they can use past and forecasted values
of the endogenous time series, past errors/shocks for the endogenous time series, and
past values of exogenous time series.
In addition, differencing of the endogenous variable may be used to improve
its stationarity, and furthermore, seasonal/periodic patterns can be utilized.
As an aside, many machine learning papers have compared their models
to ARIMA, yet SARIMAX is still efficient and is often superior to ARIMA.
In addition, the M4 and M5 Competitions indicated that Exponential Smoothing
can provide simple and accurate models.

The most straightforward time series model for MTS is a Vector Auto-Regressive ({\bf VAR}) model
with $p$ lags over $n$ variables ${\rm VAR} (p, n)$.
For example, a 3-lag, bi-variate VAR(3, 2) model can be useful in pandemic forecasting as
new hospitalizations and new deaths are related variables for which time series
data are maintained, i.e., $y_{t0}$ is the number of new hospitalizations at
time $t$ and $y_{t1}$ is the number of new deaths at time $t$.
The model (vector) equation may be written as follows:

\begin{equation}
{\bf y}_t ~=~ {\boldsymbol \delta} \,+\,
\Phi^{(0)} {\bf y}_{t-1} \,+\,
\Phi^{(1)} {\bf y}_{t-2} \,+\,
\Phi^{(2)} {\bf y}_{t-3} \,+\,
\boldsymbol{\epsilon}_t
\end{equation}

\noindent
where ${\boldsymbol \delta} \in \mathbb{R}^2$ is a constant vector,
$\Phi^{(0)} \in \mathbb{R}^{2 \times 2}$ is the parameter matrix for the first lag,
$\Phi^{(1)} \in \mathbb{R}^{2 \times 2}$ is the parameter matrix for the second lag,
$\Phi^{(2)} \in \mathbb{R}^{2 \times 2}$ is the parameter matrix for the third lag, and
${\boldsymbol \epsilon_t} \in \mathbb{R}^2$ is the residual/error/shock vector.
Some research has explored VARMA but has found them to be only slightly
more accurate than VAR models, but more complex,
as they can weigh in past errors/shocks \cite{athanasopoulos2008varma}.
SARIMAX and VAR can both be considered models for multivariate time series,
the difference is that SARIMAX focuses on one principal variable,
with the other being used as indicators for those variables,
for example, using new\_cases and hospitalizations time series to
help forecast new\_deaths.
SARIMAX tends to suffer less from the compounding of errors than VAR.
A SARIMAX model can be trimmed down to an {\bf ARX} model (Auto-Regressive, eXogenous)
to see the essential structure of the model
consisting of the first $p$ lags of the endogenous variable along with
lags $[a, b]$ of the exogenous variable ${\rm ARX} (p, [a, b])$.
For example, the model equation for an ARX(3, [2, 3]) may be written as follows:

\begin{equation}
y_t ~=~ \delta \,+\,
\phi_0 y_{t-1} \,+\,
\phi_1 y_{t-2} \,+\,
\phi_2 y_{t-3} \,+\,
\beta_0 x_{t-2} \,+\,
\beta_1 x_{t-3} \,+\, \epsilon_t
\end{equation}

\noindent
where $\hat{y}_t$, the forecasted value, is the same formula without $\epsilon_t$.
The MTS in this case consists of one endogenous time series $y_t$ and
one exogenous time series $x_t$ for $t \in 0 .. m-1$.
All the parameters $\delta, \phi_0, \phi_1, \phi_2, \beta_0$ and $\beta_1$ are now scalars.
ARX models may have more than one exogenous times series (see \cite{pearl2011exogeneity}
for a discussion of endogenous vs. exogenous variables).
The specification of a {\bf SARIMAX} model subsumes the ARX specification,

\begin{equation}
{\rm SARIMAX} (p, d, q)_{\times} (P, D, Q)_s \, [a, b]
\end{equation}

\noindent
where $p$ is the number of Auto-Regressive (AR) terms/lagged endogenous values,
$d$ is the number of stride-1 differences (Integrations (I)) to take,
$q$ is the number of moving average (MA) terms/lagged shocks,
$P$ is the number of seasonal (stride-$s$) Auto-Regressive (AR) terms/lagged endogenous values,
$D$ is the number of stride-$s$ differences to take,
$Q$ is the number of seasonal (stride-$s$) moving average (MA) terms/lagged shocks,
$s$ is the seasonal period (e.g., week, month, or whatever time period best captures the pattern) and
$[a, b]$ is the range of exogenous (X) lags to include.
Again, there may be multiple exogenous variables.
Although the optimal values for parameters of ARX models may be found
very efficiently by solving a system of linear equations, for example, using matrix factorization,
inclusion of the MA terms makes the equation non-linear, so the loss function
(e.g., Mean Squared Error (MSE) or Negative Log-Likelihood (NLL))
is usually minimized using a non-linear optimizer such as the
Limited-memory Broyden-Fletcher-Goldfarb-Shanno (L-BFGS) algorithm \cite{liu1989limited}.

\subsubsection{Second Wave}

The story is not complete for the second wave.
The M5 Competition showed the value of LightGBM.
Studies have shown that LSTM and GRU tend to perform similarly and usually
outperform Feedforward Neural Networks (FNN).

The idea of a Recurrent Neural Network ({\bf RNN}) is to take a weighted combination of
an input vector ${\bf x}_t$ (recent history) and
a hidden state vector (decaying long-term history) and
activate it to compute the new state ${\bf h}_t$.
A final layer converts this to a time series forecast(s) $\hat{y}_t$.

\begin{align*}
{\bf x}_t & ~=~ [ y_{t-p}, \dots , y_{t-1} ]                                           & {\rm input~vector} \\
{\bf h}_t & ~=~ {\bf f}(U {\bf x}_t + W {\bf h}_{t-1} \,+\, {\boldsymbol \beta^{(h)}}) & {\rm hidden~state~vector} \\
\hat{y}_t & ~=~ g (V {\bf h}_t \,+\, \beta^{(y)})                                      & {\rm output~scalar}
\end{align*}

\noindent
where $U$, $W$, and $V$ are learnable weight matrices, the $\beta$s are biases, and the variables are

\begin{itemize}
\item
${\bf x}_t \in \mathbb{R}^p$ holds the collected inputs (time series values from the recent past)
\item
${\bf h}_t \in \mathbb{R}^{n_h}$ holds the current hidden state
\item
$\hat{y}_t \in \mathbb{R}^k$ holds the 1-step to $k$-steps ahead forecasts (here $k = 1$)
\end{itemize}

\noindent
By adding sigmoid-activated gates, the flow of historical information (and thus the stability
of gradients) can often be improved.
A Gated Recurrent Unit ({\bf GRU}) model adds two gates,
while a Long Short-Term Memory ({\bf LSTM}) model adds three gates to an RNN.
Removing $W {\bf h}_{t-1}$ from the second equation turns an RNN into a {\bf FNN}.
Adding additional layers (up to a point) tends to improve forecasting accuracy.

Substantial gains seem to come from adopting an {\bf Encoder-Decoder} architecture
(used for Seq2Seq problems) where the encoder can concentrate on learning patterns
from the past, while the decoder concentrates on making accurate forecasts.
Originally, this architecture used for forecasting had LSTM/GRU units
in both the encoder and decoder.
Each step of the encoder produced a hidden state vector, with the last
one being fed into the decoder.
For a long time series, feeding all hidden state vectors
would be unlikely to help.
What was needed was to weigh these hidden states by their importance
in making forecasts, a tall order to figure out a priori,
but the weights could be learned during the training process.
This led to the development of {\it attention mechanisms} that opened
the door for more substantial improvement in time series forecasting.

Self-attention, multi-head attention, cross attention along with
positional encoding has led to the replacement of LSTM/GRU units with
encoder (and decoder) layers that use attention followed by feed-forward neural network layers.
Using several such layers for the encoder and several for the decoder,
forms a {\bf Transformer}.
In general, a Transformer consists of multiple encoder and decoder blocks.
A simplified first encoder block is depicted in Figure \ref{fig:transformer_encoder}, where
input vectors $[ {\bf x}_t ]$ are passed into a self-attention layer,
to which the input is added (via a skip connection) and then normalized to obtain $[ {\bf z}_t ]$,
after which the result is passed to a FNN, followed by another round of add-back
and normalization (see \cite{vaswani2017attention} for a more complete diagram).
In addition, as self-attention does not consider order, one may use (combine with input)
positional (based on the absolute or relative order of ${\bf x}_t$)
and/or temporal (based on ${\bf x}_t$'s date-time) encodings to make up for this.

\begin {center}
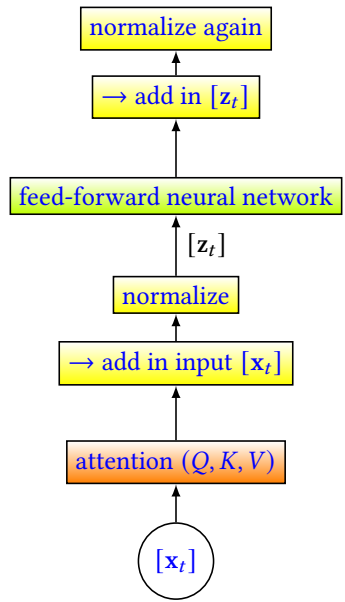
\begin{figure}[ht]
\centering
\begin {tikzpicture}[-latex, auto, node distance=1.8 cm and 1.8 cm, on grid, semithick,
state/.style ={rectangle, top color=white, bottom color=yellow, draw, text=blue, minimum width=1 cm}]

\node[state, circle, bottom color=white] (x) {$[ {\bf x}_t ]$};

\node[state, bottom color=orange] (att) [above =of x, yshift=-0.5cm] {attention $(Q, K, V)$};

\node[state] (add) [above =of att, yshift=-0.5cm] {$\to$ add in input $[ {\bf x}_t ]$};
\node[state] (norm) [above =of add, yshift=-0.9cm] {normalize};

\node[state, bottom color=lime] (fnn) [above =of norm, yshift=-0.5cm] {feed-forward neural network};

\node[state] (add2) [above =of fnn, yshift=-0.5cm] {$\to$ add in $[ {\bf z}_t ]$};
\node[state] (norm2) [above =of add2, yshift=-0.9cm] {normalize again};

\path (x) edge [left] node[below] {} (att);
\path (att) edge [left] node[below] {} (add);
\path (add) edge [left] node[below] {} (norm);

\path (norm) edge [left] node[right] {$[ {\bf z}_t ]$} (fnn);

\path (fnn) edge [left] node[below] {} (add2);
\path (add2) edge [left] node[below] {} (norm2);

\end{tikzpicture}
\caption{Transformer First Encoder Layer for a Single Head}
\label{fig:transformer_encoder}
\end{figure}
\end{center}

For some time series datasets, transformers have been shown to be the top performers.
In the related prediction problem, forecasting the next word in
Natural Language Processing (NLP), the present day Large Language Models (LLM)
based on the Transformer architecture, represent a major breakthrough.
Will the same happen for time series forecasting or will the progress
be more incremental (remember due to the stochastic nature of time series,
e.g., pandemic data, there are limits to what can be known).
Nevertheless, progress in NLP can be adapted for time series forecasting.

A less explored line of research involves
Temporal Convolutional Networks (TCN) that utilize causal and dilated convolutions,
to provide an expanded view of history,
as well as skip connections/residual blocks to utilize information from
prior layers/maintain gradients \cite{lea2016temporal}.
Although, \cite{bai2018empirical} shows advantages for TCN over LSTM for many sequence tasks,
the evidence appears to be less clear for time series forecasting.

\section{Recent Progress on Transformers for Time Series}

\subsection{Sparse Attention}

Over the last few years, there have been several papers that have
examined sparse attention for transformers.
Rather than having each time point compared with every other time
(quadratic attention), the focus is sharpened and the complexity
of the attention is reduced \cite{wen2022transformers, lin2022survey, zhu2023time}.

Given a query matrix $Q$, key matrix $K$ and value matrix $V$, attention
is computed as follows:

\begin{equation}
{\rm attention} (Q, K, V)  ~=~ {\rm softmax} \left[ \dfrac{ Q K^{^\intercal}}{\sqrt{d_k}} \right] V
\end{equation}

At a high level, sparse attention can be achieved by reducing the number of queries, or
given a query, reduce the number of keys it is compared to
(i.e., attention scores/weights computed).
Query prototypes can stand in for several queries and thereby reduce the computation.
Also, if two time points are distant from each other, setting
their attention weight to zero, would be one way to narrow the focus.
In detail, there are many ways to do this \cite{lin2022survey}.
Some of the more popular approaches are listed in Table \ref{tab:ts_papers_tr}.
Note that narrowing the focus, besides reducing computation time,
may result in improved forecasts (less distraction).

\begin{small}
\begin{table}[htbp]
\caption{Types of Time-Series Forecasting Transformer Models}
\setlength{\tabcolsep}{0.5pt}
\begin{center}
\begin{tabular}{|c|c|c|c|} \hline
{\bf Model Type} & {\bf Short Description} & {\bf Date} & {\bf Reference} \\ \hline \hline
LogTrans    & Local and LogSparse Attention & 2019 & \cite{li2019enhancing} \\ \hline
Reformer    & Only Similar Queries and Keys Are Compared & 2020 & \cite{kitaev2020reformer} \\ \hline
Informer    & Uses Selected Query Prototypes & 2021 & \cite{zhou2021informer} \\ \hline
Autoformer  & Replaces Self-Attention with Auto-Correlation & 2021 & \cite{wu2021autoformer} \\ \hline
Pyraformer  & Hierarchical/Pyramidal Attention & 2021 & \cite{liu2021pyraformer} \\ \hline
FEDformer   & \makecell{Series Decomposition and \\Use of Frequency Domain} & 2022 & \cite{zhou2022fedformer} \\ \hline
Non-stationary TRF & \makecell{Series Stationarization and \\Use of De-stationary Attention} & 2022 & \cite{liu2022non} \\ \hline
Triformer   & Triangular Structure for Layer Shrinking & 2022 & \cite{cirstea2022triformer} \\ \hline
CrossFormer & Cross-channel Modeling & 2023 & \cite{zhang2022crossformer} \\ \hline
PatchTST    & Replaces timestep inputs with patches & 2023 & \cite{nie2022time} \\ \hline
\end{tabular}
\label{tab:ts_papers_tr}
\end{center}
\end{table}
\end{small}

Due to having multiple heads and multiple layers/blocks, transformer explainability/interpretability
is challenging \cite{Chefer2021transformer}.
To a certain extent,
attention weights can also be used for interpretability \cite{lim2021time, mrini2019rethinking}.
Related research that can improve explainability/interpretability
as well as reduce training time is on simplifying transformer blocks
\cite{he2023simplifying}.

\subsection{Masking and Pre-Training}

There is an open question of how well pre-trained transformers will work
for time series forecasting.
The success of pre-training for NLP and Computer Vision (CV) problems is remarkable,
but will it carry over for MTS forecasting?

There are ways this could be applied to pandemics.
For example, transformers trained on past pandemics as well as influenza,
could be useful in future pandemics and avoid the problem
of only becoming proficient once the pandemic is past its worst peaks.

Using vectors created from words as input tokens to a transformer,
is not the same as taking a single value from a time series,
so patches may be used as they can capture a meaningful pattern
from a sub-sequence of a time series.
Given a univariate time series $[ y_t : t \in 0, \dots m-1 ]$,
it can be divided into (possibly overlapping) sub-sequences of length $p$
with a stride of $s$ (when $s = p$ there is no overlap).

\begin{equation}
{\rm patch} (p, s) ~=~ \{ [ y_t : t \in s k, \dots s k + p ] : k \in 0, \dots m / s \}
\end{equation}

\noindent
PatchTST \cite{nie2022time} takes multivariate time series data and splits it into
multiple univariate time series, considering them as independent channels. 
For each univariate time series, it creates patches, e.g., of size $p = 16$ with stride $s = 8$.
These patches are fed as tokens into a transformer.
The patches carry local semantic information and the patching reduces computation and memory usage
while attending a longer history.
Channel-independence allows for different attention maps in different channels,
promoting better handling of diverse temporal patterns.
The results show that PatchTST outperforms state-of-the-art Transformer models in
supervised and unsupervised learning tasks, including transfer learning.
Notably, it excels in forecasting with longer look-back windows. 
AR-Transformer \cite{aldosari2023transformer} also exhibits improved performance by
combining the Vanilla Transformer architecture with
segment-based attention,
teacher-forcing,
both temporal and positional encoding, and
auto-regressive (recursive) multi-horizon forecasting.

The papers \cite{tang2022mtsmae} (MTSMAE),
\cite{zha2022time} (ExtraMAE), and
\cite{li2023ti} (Ti-MAE)
discuss the use of Masked AutoEncoders (MAE) for multivariate time series forecasting.  
Some of the input patches are masked out from the encoder and the task is to train the model to essentially put them back.
Clearly, this is harder than a regular AutoEncoder that maps an input signal to a latent representation (encoded)
from which the signal is recovered (decoded).
To succeed in recovering the patches, the MAE needs to more fully capture the temporal dependencies.
With this enhanced capability, their forecasting ability, in principle, should be improved.
\cite{tang2022mtsmae} highlights the challenges in applying MAE to time series data and proposes a modified approach called MTSMAE.
The MTSMAE model incorporates the concept of Vision Transformer and leverages patching to improve
feature extraction and reduce redundancy.
In the pre-training phase, random patches from the input are masked, and the missing patches are recovered.
In the fine-tuning phase, the encoder trained in the previous step is used, and the input of the decoder is redesigned.
In contrast to the Bidirectional Encoder Representations from Transformer (BERT) decoder,
which consists of a single fully connected layer, the MTSMAE employs distinct lightweight
decoder levels based on different multivariate time series data types.
When testing this approach on various typical multivariate time series datasets
(Electricity Consuming Load, Electricity Transformer Temperature, Weather)
from diverse domains and with varying characteristics, their experimental findings indicate significantly strong performance.

Two major challenges to the successful application of the pre-training, fine-tuning
paradigm to time series forecasting are (1) the shortness of most time series,
(2) the commonness of domain/distributional shifts.
The authors of \cite{jin2022domain} propose to use domain adaptation techniques combined
with self-attention to deal with these challenges.
Other approaches include maintaining a collection of time series
from which whole series or segments are compared for similarity,
either at the raw-series level or at the representation level (see the next section).

\subsection{Representation Learning}

Rather than working with the raw multivariate time series $[ {\bf y}_t ]$,
representation learning can be used to transform the series into $[ {\bf z}_t ]$
that exist in a latent space (possibly higher dimensional) such that $[ {\bf z}_t ]$
captures the essential information in the time series in a manner that facilitates
a task like classification or forecasting.
Representation learning can be viewed as a generalization of
factor analysis \cite{xie2020representation}, and
``a good representation is one that disentangles the underlying factors of variation"
\cite{bengio2013representation}.
For forecasting, the general idea is to divide the time series into the past (up to time $\tau$)
and future (after time $\tau$) and use a function ${\bf f}$ (with parameters $W$)
to encode $Y$ and $Y^{\prime}$ into $Z$ and $Z^{\prime}$, that serve as their
richer latent representations.

\begin{align}
Y          &~=~ [ {\bf y}_1, {\bf y}_2, \dots {\bf y}_{\tau} ]          & {\rm past} \\
Y^{\prime} &~=~ [ {\bf y}_{\tau+1}, {\bf y}_{\tau+2}, \dots {\bf y}_m ] & {\rm future} \\
Z          &~=~ {\bf f}(Y, W)          & {\rm encoded ~ past} \\
Z^{\prime} &~=~ {\bf f}(Y^{\prime}, W) & {\rm encoded ~ future}
\end{align}

\noindent
The goal is to minimize the difference (measured by a loss function)
between $Y^{\prime}$ and ${\bf g} (Z^{\prime}, U)$ where ${\bf g}$
is a prediction function/network (with parameters $U$).
${\bf f}$ plays the role of an encoder function, while ${\bf g}$ plays
the role of the decoder function.
Both may have multiple layers, although some research uses, for example,
regularized regression for the prediction function
(the thought being $[ {\bf z}_t ]$ is a rich representation
that captures enough essential information from the time series
to make prediction more straightforward).

There is also work to decompose time series to better capture this
essential information, e.g., into trend, seasonal, and local variability components
(in some papers, the latter two are combined).
Due to the complex mixing of components, techniques for disentanglement
have been developed.
From the point of view of linear models, disentanglement may be seen as
a generalization of multi-collinearity reduction \cite{la2018learning}.
One way to improve a representation of time series that is likely to reduce noise
is to project it onto a smooth function using orthogonal polynomials
(e.g., Legendre, Laguerre, Chebyshev Polynomials \cite{zhou2022film}).
Training can be enhanced by augmentation or masking.
Pretext training, for example, series reconstruction using an Auto-Encoder (AE)
or Masked Auto-Encoder (MAE) may be used as well.
More for time series classification, but also for forecasting,
contrastive learning has been used.
Contrastive learning pairs up similar segments for positive cases,
and dissimilar segments for negative cases, with
the thought being prediction should be positively influenced by positive cases
and negatively by negative cases.

There are very recent studies and models developed that demonstrate the effectiveness
of representation learning for MTS as reviewed in \cite{meng2023unsupervised}.
Table \ref{tab:ts_papers_rl} highlights some of the recent work in this area.

\begin{small}
\begin{table}[htbp]
\caption{Representation Learning for Time-Series}
\setlength{\tabcolsep}{0.5pt}
\begin{center}
\begin{tabular}{|c|c|c|c|} \hline
{\bf Model Type} & {\bf Short Description} & {\bf Date} & {\bf Reference} \\ \hline \hline
TS2Vec & Local and LogSparse Attention & 2022 & \cite{yue2022ts2vec} \\ \hline
CoST & Only Similar Queries and Keys Are Compared & 2022 & \cite{woo2022cost} \\ \hline
FEAT & Uses Selected Query Prototypes & 2023 & \cite{kim2023feat} \\ \hline
SimTS & Replaces Self-Attention with Auto-Correlation & 2023 & \cite{zheng2023simts} \\ \hline
\end{tabular}
\label{tab:ts_papers_rl}
\end{center}
\end{table}
\end{small}

The paper ``TS2Vec: Towards Universal Representation of Time Series,"
uses contrastive learning with a loss function that combines
temporal and instance-based elements.
In the paper, ``CoST: Contrastive Learning of Disentangled Seasonal-Trend Representations for Time Series Forecasting,"
the authors ``argue that a more promising paradigm for time series forecasting,
is to first learn disentangled feature representations,
followed by a simple regression fine-tuning step," \cite{woo2022cost}.
Contrastive learning with loss functions for the time (for trend) and
frequency (for seasonal) domains are used.
A unique aspect of the paper,
``FEAT: A General Framework for Feature-Aware Multivariate Time-Series Representation Learning,"
is that the framework utilizes an encoder per variable/feature of a multivariate time series,
as the time sequence for each variable can have different characteristics that can be exploited usefully by the prediction function.
The authors state that,
``FEAT learns representation for the first time in terms of three diversified
perspectives: feature-specific patterns, feature-agnostic temporal patterns, and dependency between
multiple feature-specific and temporal information" \cite{kim2023feat}.
The authors of the paper, ``SimTS: Rethinking Contrastive Representation Learning for Time Series Forecasting,"
argue that while fine for time series classification, the contrastive learning approach
described in many of the papers may not be ideal for time series prediction.
In particular, their ``model does not use negative pairs to avoid false repulsion" \cite{zheng2023simts}.

Recent work wishes to make the encoding $[ {\bf z}_t ]$ more
interpretable \cite{zhao2023interpretation}
to increase user confidence in the model.
Although post-hoc interpretation methods can used, having the main model
be interpretable itself is the ideal (i.e., the model is highly accurate,
it makes sense, and there is an understanding of how variables/features affect the results).
For example, how effective were vaccinations in the COVID-19 Pandemic?

A few recent papers have shown good results with simpler architectures:

\begin{itemize}
\item
DLinear \cite{zeng2023transformers} combines series decomposition with a linear (regression) model.
A univariate time series $y_t$ is first decomposed into a simple moving average $s_t$ and
the remainder $r_t$.

\begin{align}
s_t &~=~ {\rm movingAverage} (y_t) \\
r_t &~=~ y_t - s_t
\end{align}

Then a linear model is applied to each part ($s_t$ and $r_t$) to make forecasts
that are combined together.

\item
TSMixer \cite{ekambaram2023tsmixer} is motivated by MLP-Mixer from computer vision
that relies on blocks of MLPs and does not use convolutions or self-attention,
making the architecture simpler and more efficient \cite{tolstikhin2021mlp}.
(A Multi-Layer Perceptron (MLP) is a fully connected FNN).
TSMixer looks for temporal and cross-variable dependencies in an interleaved fashion.
This is less complex than considering temporal and cross-variable dependencies
simultaneously, although if there is a strong leading indicator, this useful information
may be ignored.
\end{itemize}

\section{Recent Progress on Graph Neural Networks for Time Series}

Although Transformers are well-suited for temporal analysis,
Graph Neural Networks are conceptually well-suited for spatial-temporal analysis.
With an Encoder-Decoder or Transformer handling temporal dependencies,
a Graph Neural Network (GNN) may be more adept at capturing inter-series or spatial dependencies.

\subsection{National Level COVID-19 Data}

At the national level,
the dataset may be represented as a matrix $Y = [ y_{tj} ]$
where $t$ is time and $j$ is the variable.
The strength of GNNs is that they can more closely model and examine
the dependencies between the multiple time series.
Following this approach, each variable's time series ${\bf y}_{:j}$
can be made into a node in the graph.
Then relationship information between the nodes could be maintained
as edge properties, for example, based on cross-correlation, mutual information, etc.
Note that the strength of the, for example, cross-correlation would depend on
the lag (typically hospitalization leads death by several days).
Furthermore, if the data are not stationary, the cross-correlation
pattern may change over time.

\subsection{State-by-State COVID-19 Data}

The spread of COVID-19 in a state can affect the neighboring states over time.
This is because people travel, trade, and socialize across states.
To predict how COVID-19 will spread in a particular state, we need to consider
how it is connected to other states.
We can represent the connections between states as a graph.
In this graph, each state is a node, and there is an edge between two nodes
if there is a significant connection between the two states.
Many existing graph neural network (GNN)-based models
\cite{panagopoulos2021transfer, wang2020using} for predicting
the spread of diseases use mobility data or social connections to connect different regions and capture the spatial relationships between them.
For example, if there is a rapidly increasing curve/peak developing
in the state of New York, surely the forecasting for New Jersey
could benefit from having this information.

In addition, the study \cite{rana2023exploring} shows that states can also be connected if they
are linearly or non-linearly dependent on each other.
They calculated the correlation and mutual information between the features of the
states and found that this approach led to promising results.
For example, the study found that the number of deaths and confirmed cases of COVID-19
in Ohio and Illinois are highly correlated.
This means that there is a strong linear relationship between the two states.
As the number of deaths and confirmed cases increases in one
state, it also tends to increase in the other state.

One of the issues making this more complex is the fifty-fold expansion
of the dataset.
The dataset may now be represented as a 3D tensor ${\bf Y} = [ y_{tjk} ]$
where $t$ is time, $j$ is the variable, and $k$ is the state.
For COVID-19 weekly data, the number of time points is approximately 200,
the number of variables before any encoding is around 10, and the number
of US states around 50 (depending on whether DC and US territories are included).
The tensor would therefore have 100,000 elements, so the
number of possible dependencies is very large.

\subsection{Types of Graph Neural Networks}

Early work was in the spectral domain that utilized Fourier transforms \cite{karagiannakos2021gnnarchitectures}.
ChebNets offered a more efficient computation involving the graph Laplacian.
A Graph Convolution Network (GCN) simplifies the calculations even more,
by directly applying the Laplacian (tends to capture graph properties well).
A graph Laplacian is computed based on the graph's adjacency matrix $A$ with $I$ for self-loops
and degree matrix $D$ ($d_i$ is in-degree of node $i$).
The hidden states of the nodes (e.g., ${\bf h}_i$ for node $i$) are updated by multiplying
a normalized graph Laplacian by their previous values and a learnable weight matrix.
A Message Passing Neural Network (MPNN) is more general in that edge features
can be included in the node update calculation where a hidden state is updated
based on a combinations of its previous value ${\bf h}_i$ and
the messages ${\bf m}_{ji}$ from its neighbors (a function of both nodes and the features 
of the connecting edge with learnable weights).
Utilizing an attention mechanism to compute attention weights $a_{ij}$, a Graph Attention Network (GAT)
has the potential to better capture the dependencies between nodes.
Table \ref{tab:gnn_papers} lists six common types of GNNs with the first three being the simplest.

\begin{small}
\begin{table}[htbp]
\caption{Types of GNN Models (Sums are over Neighborhoods)}
\setlength{\tabcolsep}{0.5pt}
\begin{center}
\begin{tabular}{|c|c|c|c|} \hline
{\bf Model Type} & {\bf Short Description}   & Update for ${\bf h}_i$ & {\bf Date} \\ \hline \hline
GCN \cite{kipf2016semi}        & Graph Convolution Network       & ${\bf f}(\sum_j (d_i d_j)^{-\frac{1}{2}} W {\bf h}_j)$ & 2016 \\ \hline
MPNN \cite{gilmer2017neural}      & \makecell{Message Passing \\Neural Network}  & ${\bf f}({\bf h}_i, \sum_j {\bf m}_{ji})$ & 2017  \\ \hline
GAT \cite{velickovic2017graph}        & Graph Attention Network           & ${\bf f}(\sum_j (a_{ij} W {\bf h}_j))$ & 2017 \\ \hline
GraphSage \cite{hamilton2017inductive} & Neighborhood Sampling            & algorithm 1 in \cite{hamilton2017inductive} & 2017  \\ \hline
GIN \cite{xu2018powerful}       & Graph Isomorphism Network       & equation 4.1 in \cite{xu2018powerful} & 2018 \\ \hline
TGN \cite{rossi2020temporal}       & \makecell{Temporal Graph Network \\(dynamic) }  & embedding equation in \cite{rossi2020temporal} & 2020 \\ \hline
\end{tabular}
\label{tab:gnn_papers}
\end{center}
\end{table}
\end{small}

In utilizing GNNs for MTS forecasting, researchers have tried various ways
to define the underlying graph structure for the GNN.
A static graph $G$ is easier to deal with, but then the question
is whether it is given a priori or needs to be learned from data
(graph structure learning).
If the graph is dynamic, its topological structure changes over time.
For time series, the graph structure (nodes/edges) would typically
change a discrete points in time, i.e., the graph at time $t$, $G_t$.

Section 4 in a survey of GNNs for time series forecasting \cite{jin2023survey}
mentions two types of dependencies that models need to handle:
(1) modeling spatial or inter-variable dependencies and
(2) modeling temporal dependencies.
GNNs are ideally suited for (1), but for (2) are often combined with
recurrent, convolutional, or attention-based models.

Several studies have leveraged Graph Neural Networks (GNNs) for COVID-19 forecasting.
Kapoor et al. \cite{kapoor2020examining} used a spatial-temporal GNN to incorporate mobility data,
capturing disease dynamics at the county level.
Panagopoulos et al. \cite{panagopoulos2021transfer} introduced MPNN-TL, a GNN for understanding
COVID-19 dynamics across European countries, emphasizing the role of mobility patterns.
Fritz et al. \cite{fritz2022combining} combined GNNs with epidemiological models,
utilizing Facebook data for infection rate forecasting in German cities and districts.
Cao et al. \cite{cao2020spectral} developed StemGNN, employing the Graph Fourier Transform (GFT)
and Discrete Fourier Transform (DFT) to capture temporal correlations in a graph structure
representing different countries and forecasting confirmed cases across multiple horizons. 
temporal modeling within a single deep neural network (DNN) block,
eliminating the need for separate treatment of neighboring areas and capturing both spatial and temporal dependencies.
It also presents a flexible architectural design by concatenating multiple DNN blocks,
allowing the model to capture spatial dynamics across varying distances and longer-term temporal dependencies.

Combining Transformers and Graph Neural Networks can provide the advantages of both.
For example, SageFormer \cite{zhang2023sageformer} uses a graph representation and GNNs
to establish the connections between the multiple series and as such helps focus
the attention mechanism.

\section{Foundation Models}

A foundation model serves as a basis for more general problem solving.
The term foundation model appeared in \cite{bommasani2021opportunities}
with the general notion predating this. 
For example, one could view transfer learning as a precursor to foundation models.
This paper argues that even though foundation models are based on
deep learning and transfer learning, their large scale supports
broader applications and emergent capabilities, i.e., homogenization and emergence.

In forecasting, whether it be a traditional statistical model or a deep learning
model, the main idea is to train the model for a particular dataset,
so that it can pick up its specific patterns.
Unfortunately, in many cases, there are not enough data available to train
a complex model that has many trainable parameters.
Data augmentation techniques \cite{javeri2021improving} can help on the margins.
At the beginning of pandemics, the problem is severe.
Just when the need the accurate forecasts is the greatest,
the amount of data is inadequate.
This is where foundation models can show their value.

Foundation models with billions of parameters have recently shown
remarkable success in the areas of natural language and computer vision.
Naturally, other domains are investigating how foundation models can
work for other modes of data as well as for multi-modal data.

As time series is a type of sequential data, as is natural language,
one might expect foundation models for time series to work as well as
Large Language Models (LLMs) do for natural language.
A foundation model, having been extensively trained,
should more readily pick up the pattern in a new dataset.
There are several reasons why the problem is more challenging
in the times series domain:

\begin{itemize}
\item
Variety.
Time Series data are being collected for many domains,
many of which will have their own unique characteristics.
The patterns seen in stock market data will be much
different from Electrocardiogram (EKG/ECG) signals.
\item
Many Small Datasets.
Before the era of big data, time series data mainly consisted
of short sequences of data and this makes it hard for deep
learning models to gain traction.
This characteristic of time series data will remain to a lesser degree
over time.
\item
Lack of Lexicon, Grammar, and Semantics.
Any sequence of numbers can form a time series.
This is not the case with natural language as only certain
sequences of lexical units are meaningful,
i.e., there is more structure to the patterns.
Although time series may be decomposed into trend, seasonal,
and local patterns, the structural restrictions are not comparable.
\end{itemize}

\subsection{Backbone Model/Architecture}

A foundation model can be built by scaling up a (or a combination of) deep learning model(s).
Exactly how this is done is the secret sauce of today's highly successful Large Language Models (LLMs),
such as GPT, BART, T5, LLaMA, PaLM, and Gemini \cite{wan2023efficient}.
A comparison of two recent multimodal LLMs, OpenAi's Chat-GPT4 and Google's,
is given in \cite{mcintosh2023google}.
The efficiency of LLMs is discussed in \cite{wan2023efficient}.

Multiple backbone models/architectures have been considered for foundation models
in time series classification \cite{yeh2023toward}.
This paper compared LSTM, ResNet, GRU, and Transformers architectures with the Transformer
architecture showing the most promise.
For multivariate time series, models with focused or sparse attention have
shown greater accuracy (smaller errors) than transformers using full attention. 
Furthermore, transformer-based backbone models may follow the encoder-decoder architecture
or may replace either the encoder or decoder with simpler components.
The debate now mainly centers around whether to use an encoder-decoder or decoder-only
architecture \cite{fu2023decoder}.
Both types of transformers are represented by current state-of-the-art LLMs:
Generative Pre-trained Transformer (GPT) is decoder-only, while
Bidirectional and Auto-Regressive Transformers (BART) and
Test-to-Text Transfer Transformer (T5) are encoder-decoders.

There are additional architectures that may be considered for a backbone:
(1) Transformer++ architecture extends self-attention to include convolution-based heads 
that allow tokens/words to be compared with context vectors representing multiple tokens
\cite{Thapak2020transformer++}.
Half the heads use scaled-dot product attention with the other half using convolutional attention.
This allows additional temporal or semantic dependencies to be captured.
(2) State-Space Models,
for example, Mamba combines elements from MLPs (of Transformers), CNNs, and RNNs,
as well as classical state space models \cite{gu2023mamba}
to provide a more efficient alternative to plain transformers.

Another issue relevant to time series is whether to use
channel-independent or cross-channel modeling \cite{han2023capacity}.
As discussed, PatchTST successfully utilized channel-independence.
One form of cross-channel modeling would be to consider cross-correlations
between different channels/variables (e.g., hospitalizations and new deaths).
As shown in Figure \ref{fig:dailydeath}, results are likely to be better if lagged
cross-correlations are used.
There is also an issue of which variables/channels to include as
including many variables may be counter productive (analogs to
the problem of feature selection in VAR and SARIMAX models).

For the spatial-temporal domain, the current research on Graph Foundation Models (GFMs)
\cite{liu2023towards} becomes more relevant.
Foundation models are typically scaled-up transformers and work well for sequence
data such as natural language and time series.
Other types of deep learning models may be useful for data having a spatial component.
Convolution Neural Networks and Graph Neural Networks intuitively match this type of data.
Two types of architectures serve as the most popular backbones:
Message-passing-based GNNs \cite{gilmer2017neural} and transformer-based \cite{ying2021transformers}.
Several studies adopt GAT, GCN, GIN, and GraphSage as their backbone architectures with GIN
being particularly favored due to its high expressive power.
These GNNs are often integrated with RNNs to capture temporal dependencies within the data,
and they can be scaled up to form a foundation model.

The success of transformers has given rise to the second type of backbone which is a hybrid
of transformers and GNNs.
This method improves upon traditional message-passing GNNs by having strong expressive power and
the ability to model long-range dependencies effectively.
GROVER \cite{rong2020self} uses a GNN architecture to capture the structural
information of a molecular graph, which produces the outputs in the form of queries, keys,
and values for the Transformer encoder.
For a heterogeneous graph, researchers commonly employ the Heterogeneous Graph Transformer (HGT)
\cite{hu2020heterogeneous} as the encoder.


\subsection{Building a Foundation Model for Time Series}

There are at least four approaches for creating foundation models for time series:

\begin{enumerate}
\item
Use the power of an existing Large Language Model.
This would involve converting time series segments or patches to words,
using these to generate the words that follow and then converting
back to time series (i.e., the forecasts).
The basis for this working would be the existence of universal patterns
in the two sequences (words and time series segments).
However, without care, the time series converted to a sequence of
words are likely to produce meaningless sentences.
The same might happen when the output words are converted to the time series forecast.
Fine-tuning a Large Language Model using time series data may improve
their forecasting capabilities.
\item
Build a general-purpose Foundation Model for Time Series from scratch using a huge number of
time series datasets.
This would be a large undertaking to collect and pre-process the large volume
of time series data.
High performance computing would also be needed for extensive training.
Although comprehensive training in the time series domain is generally considered
to be less demanding than the language domain.
\item
Build a special purpose Foundation Model for Time Series from scratch using
datasets related to disease progression.
This alternative is more manageable in terms of the volume of training data needed
and the training resource requirements.
Also, it is unknown whether there exists exploitable universality across time series domains.
Would a foundation model trained on stock market data be useful for pandemic prediction?
\item
Create a Multi-modal Foundational Model that contains textual and time series data.
For example, the text could be from news articles or social media about the COVID-19 pandemic and
the time series data (weekly/daily) could be from the CDC or OWID.
The two would need to be synchronized based on timestamps and using
techniques such as Dynamic Time Warping (DTW) for time series alignment \cite{muller2007dynamic}.
\end{enumerate}

Very recently, there have been several efforts to create foundation models for time series forecasting,
as indicated in Table \ref{tab:foundation_papers}.
The model type indicates which of the above four approaches are taken.
The backbone indicates the base deep learning technique from the foundation model
is built up.

\begin{small}
\begin{table}[htbp]
\caption{Foundation Models for Time Series}
\setlength{\tabcolsep}{0.5pt}
\begin{center}
\begin{tabular}{|c|c|c|c|c|} \hline
{\bf Model Type} & {\bf Model} & {\bf Short Description} & {\bf Backbone} & {\bf Date} \\ \hline \hline
2 & TimeCLR    \cite{yang2022timeclr, yeh2023toward}  & Contrastive Learning pre-training & Transformer & Jun 2022 \\ \hline
2 & TimesNet   \cite{wu2022timesnet}        & 2D variation modeling & TimeBlock  & Oct 2022 \\ \hline
1 & GPT4TS     \cite{zhou2023one}           & LLM with patch token input & GPT-2 & May 2023 \\ \hline
1 & LLM4TS     \cite{chang2023llm4ts}       & Temporal Encoding with LLM & GPT-2 & Aug 2023 \\ \hline
4 & UniTime    \cite{liu2023unitime}        & Input domain instructions + time series & GPT-2  & Oct 2023 \\ \hline
2 & TimeGPT    \cite{garza2023timegpt}      & Added linear layer for forecasting & Transformer & Oct 2023 \\ \hline
4 & Time-LLM   \cite{jin2023time}           & Input context via prompt prefix & LLaMA & Oct 2023 \\ \hline
2 & PreDct     \cite{das2023decoder}        & Patch-based Decoder-only & Transformer & Oct 2023 \\ \hline
2 & Lag-Llama  \cite{rasul2023lag}          & Lag-based Decoder-only & LLaMA & Oct 2023 \\ \hline
3 & AutoMixer  \cite{palaskar2023automixer} & Adds AutoEncoder to TSMixer & TSMixer & Oct 2023 \\ \hline
1 & TEMPO      \cite{cao2023tempo}          & Pre-trained transformer + statistical analysis & GPT-2 & Oct 2023 \\ \hline
1 & PromptCast \cite{xue2023promptcast}     & Text-like prompts for time series with LLM & GPT-3.5 & Dec 2023 \\ \hline
\end{tabular}
\label{tab:foundation_papers}
\end{center}
\end{table}
\end{small}

\subsubsection{\bf Type 1: Repurposed LLMs}

This type utilizes large language foundation models and repurposes them for time series data.
The LLM architecture designed for textual data is suited for time series due to its sequential nature.
Due to being pre-trained on large datasets using billions of parameters, it has shown satisfactory
results with fine-tuning for specific language processing tasks such as question answering,
recommendation, and others.
In addition, these can also be fine-tuned for time series forecasting tasks related to disease,
weather, and energy consumption forecasting.
However, the transfer of pre-trained LLM to time series data has several requirements.
LLMs require tokens as inputs.
While a single point can be used as input, it cannot cover semantic information.
Therefore, most of these models divide the time series into patches of a certain length.
These patches are considered as tokens which can be used as input to an LLM.
To mitigate the distribution shift, GPT4TS \cite{zhou2023one}, LLM4TS \cite{chang2023llm4ts}
use reversible instance normalization (RevIN) \cite{kim2021reversible}.
While GPT4TS \cite{zhou2023one} uses patch tokens as input to GPT, other methods enhanced
the tokens with further encodings.
LLM4TS \cite{chang2023llm4ts} encodes temporal information with each patch and considers
the temporal details of the initial time step in each patch.
Temporal encoding of input tokens is important for domain-specific models used for disease
forecasting as data for certain viral infections like COVID-19 show rise in holiday seasons
and cold weather.
TEMPO \cite{cao2023tempo} decomposes the time series into seasonal, trend and residual components.
Additionally, it uses a shared pool of prompts representing time series characteristics.
The decomposed seasonal, residual and trend components are individually normalized, patched, and embedded.
The embedded patches are concatenated with the retrieved prompts before passing as input to a GPT block.
Different from methods tokenizing time series data, PromptCast \cite{xue2023promptcast} frames
the time series forecasting as a question answering task and represents the numerical values as sentences.
It uses specific prompting templates to apply data-to-text transformations of time series to sentences.

The models GPT4TS \cite{zhou2023one}, LLM4TS \cite{chang2023llm4ts}, TEMPO \cite{ cao2023tempo}
use GPT as their backbone network, which is decoder only.
However, these models are non-autoregressive and use a flattened and linear head on the last raw
hidden state from the decoder to estimate chances for likely outcomes in a horizon window.
However, this does not allow forecasting with varying horizon lengths which could be important for
making timely decisions related to pandemic and epidemic control.

\subsubsection{\bf Type 2: Broadly Pre-trained on Time Series Datasets}

This type designs a model specifically targeted for time series data and utilizes pre-training from scratch.
TimeGPT \cite{garza2023timegpt} is among the initial foundation models that use an encoder-decoder
architecture with multiple layers containing residual connections and layer normalization.
The linear layer maps the output of the decoder to a horizon window to estimate the likely outcome.

As a foundation model, TimeCLR \cite{yeh2023toward} was pre-trained for time series classification.
TimeCLR utilizes self-supervised learning for pre-training with a transformer backbone.
The method adopts a contrastive learning pre-training method, building on the existing method, SimCLR.
Several time series data augmentation techniques were utilized including jittering, smoothing, magnitude warping, time warping,
circular shifting, adding slope, adding spikes, adding steps, masking, and cropping.
These additional techniques enhance the pre-training process by allowing the model to learn more invariance properties.
The overall architecture consists of a backbone and projector.
The backbone has a transformer architecture and the projector consists of linear and ReLU layers.
For fine-tuning, a classifier model was added on top of the projector, and the backbone, projector, and classifier model
were updated using cross-entropy loss.

Another pre-trained foundation model, PreDct \cite{das2023decoder} utilizes a decoder-only architecture
in an autoregressive manner to allow estimation of likely outcomes for varying history lengths, horizon
lengths and temporal granularities.
The input is pre-processed into patches using residual layers.
The processed patches are added to positional encoding before being passed as input to the network.
The network consists of a stack of transformer layers where each transformer layer contains self-attention
followed by a feed-forward network.
It uses causal self-attention, that is each token can only attend to tokens that come before it and trains
in decoder-only mode.
Therefore, each output patch token only estimates for the time period following the last input patch
corresponding to it.
While these methods use a certain patch length, different patch lengths may be optimal for
different time series data.
For example, COVID-19 and other viral infections show periodicity over extended time periods, whereas,
weather data demonstrate daily variations.
To identify this periodicity, TimesNet \cite{wu2022timesnet} uses the Fast Fourier Transform to discover
the optimal periods and stack these in a column-wise manner to represent the input in a 2D format.
In the 2D format the inter-period and intra-period features may be estimated using 2D kernels/filters from
existing developed models such as ConvNext, CNN, DenseNet, and others.

Another model, Lag-Llama \cite{rasul2023lag} is built using LLaMA that features accuracy with a reduced
number of parameters, normalization with RMSNorm \cite{ zhang2019root} and RoPE \cite{su2024roformer}
as well as SwiGLU \cite{shazeer2020glu} activation and optimization using AdamW \cite{touvron2023llama}.
(Root Mean Squared Norm (RMSNorm) divides by the RMS and serves as a faster alternative to LayerNorm
that subtracts the mean and divides by the standard deviation.
Rotary Positional Embedding (RoPE) is a form for relative position encoding (used for self-attention)
that encodes the absolute position with a rotation matrix and incorporates the explicit relative position
dependency in self-attention formulation.
SwiGLU is a smoother replacement for the ReLU activation function that combines Swish ($x {\rm sigmoid} (x)$)
with Gated Linear Units.
AdamW improves upon the Adam optimization algorithm by utilizing decoupled weight decay regularization.)
The inputs are based on selected lags that may include, for example, seasonal effects.
This makes it different from using patches.
It performs probabilistic forecasting to estimate a distribution for the next value(s).
Alternatively, one could focus on point and/or interval forecasting.
A current limitation is that it only works on univariate time series
(further innovations will be needed for multivariate time series).
Additionally, the longer context window reduces efficiency and increases memory usage due to
increased evaluations in attention modules.
However, the longer context window enables models to process more information, which is particularly
useful for supporting longer histories in case of diseases like influenza.
A follow-up version, Llama 2 \cite{touvron2023llama2}, works towards improving efficiency.
It is mostly like Llama 1 in terms of architecture and uses the standard transformer architecture
and uses pre-normalization using RMSNorm, SwiGLU activation function, and rotary positional embeddings.
Its two primary differences include doubling the context length and using grouped-query attention
(GQA) \cite{ainslie2023gqa} to improve inference scalability.
The context window is expanded for Llama 2 from 2048 tokens to 4096 tokens which enables models
to process more information.
For speeding up attention computation, the standard practice for autoregressive decoding is to cache
the key (K) and value (V) pairs for the previous tokens in the sequence.
With doubled context length, the memory costs associated with the KV cache size in attention models
grow significantly.
As KV cache size becomes a bottleneck, key and value projections can be shared across attention heads
without much degradation of performance.
For this purpose, a grouped-query attention variant with 8 KV projections is used.
Grouped-query attention (GQA), a generalization of multi-query attention, uses an intermediate
(more than one, less than the number of query heads) number of key-value heads to reduce memory usage.

\subsubsection{\bf Type 3: Pre-trained on Domain-related Time Series Datasets}

This type of model is pre-trained on domain-related data.
Each domain has different characteristics related to seasonality and trend.
For example, pandemic-related data show an increasing trend in the initial days of a disease outbreak,
while energy consumption fluctuates greatly within a year.
Therefore, pre-training on a specific domain may provide improved performance.
Among special purpose foundation models, AutoMixer \cite{palaskar2023automixer} was trained
for business and IT observability.
Different from other foundation models, it poses channel compression as a pre-training task.
It proposes to project the raw channels to compressed channel space where unimportant channels
are pruned away, and only important correlations are compactly retained.
RNN-based AutoEncoder (AE) handling variable input and output sequence lengths is used for pre-training.
For fine-tuning, the input is compressed using the encoder part of pre-trained AE.
Afterwards, the compressed representation is passed to TSMixer which is trained from scratch.
The output from TSMixer is passed as input to the decoder part of the AE to get the results
for the horizon window.

\subsubsection{\bf Type 4: Multimodal with Text and Time Series}

Previous work has shown how information from news or social media can improve forecasting
\cite{pai2018predicting}.
Most of this work required feature engineering, for example, using sentiment analysis scores
to improve sales or stock market predictions.
However, multi-modal foundation models provide greater potential and increased automation.

Type 4 models utilize both textual and time series data to improve forecasting accuracy and provide
greater potential for explainability.
In the case of pandemic estimates, a model trained on both disease outbreaks and additional
textual information about vaccination development may enhance the results for future diseases.
Time-LLM \cite{jin2023time} introduces a multimodal framework utilizing a pre-trained large language
foundation model.
The input time series was tokenized and embedded via patching and a customized embedding layer.
These patch embeddings are then reprogrammed with condensed text prototypes to align two modalities.
Additional prompt prefixes in textual format representing information about input statistics are
concatenated with the time series patches.
The output patches from the LLM are projected to generate the forecasts.

UniTime \cite{liu2023unitime} allows the use of domain instructions to offer explicit domain identification information to the model.
This facilitates the model to utilize the source of each time series and adapt its forecasting strategy accordingly.
Specifically, it takes the text information as input and tokenizes it as done in most language processing models.
In addition, it masks the input time series to alleviate the over-fitting problem.
A binary indicator series is formed representing the masked and unmasked locations.
Both the masked time series and binary indicator series are tokenized through patching, embedded into a hidden space,
and fused through gated fusion (for optimized mixing of signals).
The fused patch tokens and the text tokens are concatenated before passing as input to the language model decoder.
The output tokens from the language model decoder are padded to a fixed sequence length.
The padded result is then passed through a lightweight transformer.
To allow variable horizon lengths, a linear layer is afterward utilized with a maximum length parameter to generate predictions.
The model always outputs that number of values, which may be truncated to get estimates for a certain horizon window.

\subsection{Pre-Training Foundation Models}

As pointed out in \cite{godahewa2021monash}, there is a paradigm shift in
time series forecasting from having a model trained for each dataset,
to having a model that is useful for several datasets.
This paradigm shift naturally leads to foundation models
for time series forecasting that are trained over a large number of datasets.
As discussed in the next subsection, the accuracy of such models
can be improved by fine-tuning. 

Pre-training of a foundation model for time series is challenging due to the
diversity of data, but easier in the sense that the volume and dimensionality are less than for LLMs.
Finding enough datasets is another problem to deal with.
As a partial solution, the following repositories include time series datasets from multiple domains:

\begin{itemize}
\item
The Monash Time Series Forecasting Repository \cite{godahewa2021monash}
contains 30 dataset collections many of which contain a large number of time series
(summing to a total of 414,502 time series).
\url{https://forecastingdata.org/}

\item
The University of California, Riverside (UCR) Time Series Classification Archive \cite{dau2019ucr}
contains 128 datasets.
Its focus is on univariate time series classification.
\url{https://www.cs.ucr.edu/~eamonn/time_-series_-data_-2018/}

\item
University of East Anglia (UEA) Repository \cite{bagnall2018uea}
contains 30 datasets.
Its focus is on multivariate time series classification.
\end{itemize}

In the time series domain, Self-Supervised Learning \cite{zhang2023self}
can be utilized for large scale training to deal with the lack of labeled data.
As time series forecasting is not dependent on labels anyway,
how would self-supervised learning differ from standard training?
It can enhance it:
For example, as a pretext subtask, a portion of a time series may be masked out
and regenerated.
The thought being is that doing so can help the model (especially a foundation model)
make accurate forecasts.
Careful use of data augmentation may be beneficial, so long as it does
not interrupt complex temporal patterns.
Furthermore, adding and removing noise from the time series may help
the model to see the true patterns.
Of course, self-supervised learning is essential for other
time series subtasks such as time series classification and anomaly detection.
Even if forecasting is the goal, training on related subtasks is
thought to improve the overall capabilities for foundation models.

Neural Scaling Laws \cite{kaplan2020scaling} for LLMs indicate that error rates drop following 
a power law that includes training set size and number of parameters
in the model.
As these get larger, the demand for computing resources and time naturally increase.
To reduce these demands, \cite{sorscher2022beyond} has shown that by using a good self-supervised
pruning metric to reduce training samples, for example, one could obtain a drop
``from 3\% to 2\% error by only adding a few carefully chosen training examples,
rather than collecting 10× more random ones."
Exactly, how this translates to time series forecasting is still an open research problem.

\subsection{Fine-Tuning Foundation Models}

As foundation models have a large number of trainable parameters that are optimized
using a huge amount of data requiring high performance computing over extended
periods of time, how can they be generally useful for time series forecasting?

The idea of fine-tuning a foundation model is to make small adjustments
to the parameters to improve its performance for a particular sub-domain.
For example, a foundation model trained on infectious diseases could be fine-tuned
using COVID-19 datasets.

A foundation model that follows a transformer architecture has trainable parameters
for its self-attention mechanism and in its multiple fully connected layers.
One option would be to re-train the final layers of the last few blocks of
the transformer, freezing the rest of the parameters.
One may also re-train the attention weights.
The best combination to provide efficient fine tuning is an ongoing research problem.
A lower cost solution is to only re-train the attention weights.

The goal of Parameter Efficient Fine-Tuning (PEFT) \cite{houlsby2019parameter, liao2023parameter}
techniques is to increase the accuracy of the pre-trained model as efficiently as possible.
Three common approaches are listed below.

\begin{itemize}
\item
Sparse Fine-Tuning provides a means for choosing which parameters to fine-tune
based on, for example, changes in parameter values or values of gradients.
The majority of parameters remaining frozen.
\item
Adapter Fine-Tuning adds new trainable weight matrices, for example, after each feed-forward
neural network in a transformer, one ($W^{(dn)}$) to project down from the dimension of the transformer
model $d$ to a lower dimension $r$, and the other ($W^{(up)}$ restoring back to dimension $d$.
Given a hidden state vector ${\bf h}$, it will be updated to the following value.

\begin{equation}
{\bf h} \,+\, f ({\bf h} W^{(dn)}) \, W^{(up)}
\end{equation}

Fine-tuning only changes $W^{(dn)}$ and $W^{(up)}$ with all the other parameters being frozen.

\item
Low Rank Adaptation (LoRA) \cite{hu2021lora} is similar to adapter fine-tuning,
but is integrated into existing layers, e.g., given a linear layer with
computation ${\bf h} W$, $W^{(dn)}$ and $W^{(up)}$ are added as follows:

\begin{equation}
{\bf h} W \,+\, ({\bf h} W^{(dn)}) \, W^{(up)}
\end{equation}

The advantage of LoRA is through pre-computation, its inference speed is the
same as with full fine-tuning.
A limitation of this technique is that it cannot be applied to a unit having
a nonlinear (activation) function.
\end{itemize}

Three common ways of improving the accuracy of foundational models for particular domains
are Fine-Tuning (FT), Retrieval Augmented Generation (RAG), and Prompt Engineering (PE).
Used in combination the effect can be substantial \cite{gao2023retrieval},
e.g., hallucinations in LLM can be reduced and the timeliness of answers increased.
Retrieval Augmented Generation is facilitated by maintaining rapid access to continually
updated sources of information for example stored in relational databases or knowledge graphs.
Prompt Engineering supplements a query with relevant information to make the foundation
model aware of it.
RAG can be used to support building the prompts.
RAG can also help with and facilitate the fine-tuning of foundation models.

\section{Use of Knowledge}

Data-driven methods have made great strides of late, but may still benefit by
using accumulated knowledge.
Even the remarkably capable recent Large Language Models improve
their responses using knowledge.
For Pandemic Prediction, knowledge about the disease process and
induced from previous studies can improve forecasting models.
Knowledge about the future based on industrial or governmental
policies can be very useful in forecasting, e.g.,
schools or stores will open in two weeks, a mask mandate will start
next week, etc.

The use of knowledge has been a goal that has been pursued for a long
time for time series forecasting.
For example \cite{armstrong1993causal} used 99 rules based on causal forces
to select and weigh forecasts.
Then, the causal forces (growth, decay, supporting, opposing, regressing)
were specified by the analyst (but could be learned today).
Another direction for applying knowledge is situational awareness
\cite{peng2019knowledge}.
Knowledge can be useful in feature selection either for improved
forecasting or greater interpretability.
It can be used in model checking, e.g., in a pandemic the calculated
basic reproduction number $R_0$ based on the forecast is outside the feasible range.

To improve the forecasting of fashion trends, the authors of \cite{ma2020knowledge}
have developed the Knowledge Enhanced Recurrent Network (KERN) model
and shown that the incorporation of knowledge into the model has increased its
forecasting accuracy.
The base model follows an LSTM encoder-decoder architecture
to which internal knowledge and external knowledge are added.
For example, they develop close-far similarity relationships for trend patterns
as internal knowledge (alternatively could be taken as a different view of the data)
to create a regulation term to add to the loss function.

As external (or domain) knowledge, they utilize a fashion element ontology (taxonomy and part-of relationships).
Then if sales of a dress part (e.g., peplum) go up, it would be likely that the sales of dresses would go up.
This external knowledge is incorporated via the embedding of the inputs that are passed to the encoder.
The authors note the improvements due to adding knowledge, particularly for longer-term forecasting.

\subsection{COVID-19 Knowledge Graphs}

There are many large Knowledge Graphs available that are built as either Resource Description Framework (RDF)
Graphs or Labelled Property Graphs (LPG).
The RDF graphs consist of triples of the form (subject, predicate, object) and
LPG graphs can be mapped to triples.
A Temporal Knowledge Graph (TKG) may be viewed as a collection of quads $(s, p, o, t)$
meaning that predicate $p$ applied to subject $s$ and object $o$ is true at time $t$.
Intervals may be used as well $(s, p, o, [t_1, t_2])$.
Table \ref{tab:kg_papers} lists some available knowledge graphs that contain
information about COVID-19.

Several utilize CORD-19.
The Allen Institute for AI has made available a large collection of research papers
on the COVID-19 Pandemic:
COVID 2019 Open Research Dataset (CORD-19) dataset,
or in RDF,
CORD-19-on-FHIR: Linked Open Data version of CORD-19.

\begin{small}
\begin{table}[htbp]
\caption{COVID-19 Knowledge Graphs}
\setlength{\tabcolsep}{4pt}
\begin{center}
\begin{tabular}{|c|c|c|c|c|} \hline
{\bf KG}         & {\bf Type}   & {\bf Base}   & {\bf Date} \\ \hline \hline
COVID-19-Net \cite{rose2020peter}      & LPG/Neo4j    & many               & 2020 \\ \hline 
Covid-on-the-Web \cite{michel2020covid} & RDF/Virtuoso & CORD-19            & 2020  \\ \hline 
Cord19-NEKG \cite{michel2020covid}       & RDF/Virtuoso & CORD-19            & 2020 \\ \hline 
COVID-KG \cite{steenwinckel2020facilitating}        & RDF          & CORD-19            & 2020  \\ \hline
CovidGraph \cite{gutebier2022covidgraph}      & LPG/Neo4j    & many               & 2022  \\ \hline
CovidPubGraph \cite{pestryakova2022covidpubgraph}   & RDF/Virtuoso & CORD-19            & 2022  \\ \hline
COVID-Forecast-Graph \cite{zhu2022covid} & RDF/OWL  & COVID Forecast Hub & 2022  \\ \hline
\end{tabular}
\label{tab:kg_papers}
\end{center}
\end{table}
\end{small}

\noindent

So far, there is little work on building Temporal Knowledge Graphs (TKGs) for COVID-19,
even though they match up well with forecasting.
The approach taken by
Temporal GNN with Attention Propagation (T-GAP) \cite{jung2021learning}
could be used to build a TKG for COVID-19.
T-GAP performs Temporal Knowledge Graph Completion
that can fill in missing information as well as make forecasts.
The GNN uses information from the TKG based on the current query,
as well as Attention Flow
(multi-hop propagation with edge base attention scores), to make TKG completion more accurate
This approach also improves the interpretability of the model.

\subsection{Temporal Knowledge Graph Embedding}

Temporal Knowledge Graph Embedding (TKGE) can be used for link prediction and
if time is in the future, it involves forecasting.
TKGE represents graph elements in latent vector spaces with relationships
(including temporal ones) determining the relative positions of the vectors.
There is a growing list of TKG embedding techniques including,
TAE, TTransE, Know-Evolve, TA-TransE, TA-DistMult, DE-SimplE, TEE, and ATiSE,
with the last one including time series decomposition \cite{xu2019temporal}.
It is an open question as to what degree these vectors (as knowledge of
temporal relationships) could improve other deep learning forecasting models.
The possible synergy between link prediction in TKGs and MTS forecasting
needs further exploration.

\subsection{Incorporation of Knowledge}

There a several ways that knowledge can be incorporated into a deep learning model:

\begin{enumerate}
\item
Composite Loss Function:  e.g.,
$(1 - \lambda) \| {\bf y} - {\bf \hat{y}} \|_p + \lambda \| {\bf z} - {\bf \hat{y}} \|_p$
where ${\bf y}$ is the actual time series, ${\bf \hat{y}}$ are the predicted values and
${\bf z}$ are predictions from a theory-based or simulation model.
\item
Applying Constraints:  e.g.,
$\| {\bf y} - {\bf \hat{y}} \|_p + \lambda f_c({\bf \hat{y}})$
where $f_c$ is a penalty function based on constraint violation.
Depending on the form of the constraint, it could be viewed as regularization.
\item
Factored into Self-Attention Mechanism:
From previous studies, controlled experiments, or theory,
the relevance \cite{bai2022enhancing} of $y_{tj}$ to $y_{t-l,k}$ could be maintained,
for example, in a temporal knowledge graph (variable $j$ to $k$ with lag $l$) and used to focus
or modify self-attention calculations.
\item
Embedded and Combined with Input:
A sub-graph of a COVID-19 (Temporal) Knowledge Graph would produce embedded vectors that would be
combined (e.g., concatenated) with the input multivariate time series (e.g., raw or patch level).
\item
Injected into a Downstream Layer:
Determining the ideal place to combine knowledge with input data or latent representations thereof
is challenging.
For models based on representation learning that map ${\bf x}_t$ to ${\bf z}_t$,
it could happen anywhere in the process before the final representation is created.

\item
Knowledge Influencing Architecture:
A sub-graph of a COVID-19 (Temporal) Knowledge Graph could also be used as a draft architecture
for GNN.
\end{enumerate}

\subsection{Knowledge Enhanced Transformers}

Use of future knowledge is exploited by Aliformer
by modifying the transformer self-attention mechanism \cite{qi2021known}.
In the e-commerce domain, they consider two types of future knowledge:
produce related and platform-related.

There is ongoing research on the use of knowledge for the improvement
of Large Language Models.
Pre-trained Language Models (PLM) are typically large transformers
that are extensively trained and then fine-tuned, and include
BERT,
Generative Pre-trained Transformer (GPT),
Bidirectional and Auto-Regressive Transformers (BART), and
Test-to-Text Transfer Transformer (T5) \cite{min2021recent}.
These models can be enhanced with knowledge:
The survey in \cite{yang2021survey} discusses how
symbolic knowledge in the form of entity descriptions,
knowledge graphs, and rules can be used to improve PLMs.
A key question is how to design an effective knowledge
injection technique that is most suitable for a PLM's architecture.

Multivariate Time-series forecasting is a critical aspect of pandemic forecasting.
As of yet, an accurate forecasting model may not be built solely using an LLM.
Fine-tuning using pandemic literature and prompt design can help LLM improve its forecasting capability.
Still, it can be highly beneficial for another model that specializes in capturing temporal patterns
in MTS COVID-19 datasets to be applied.
The LLM can be used to improve the MTS model or be used in conjunction with it.

Just as knowledge graphs can be used to enhance PLM's performance on language tasks,
they can also be used to improve the accuracy and explainability of transformers
used for MTS forecasting.
Recent literature suggests that traditional attention may not be necessary for capturing temporal dependencies
\cite{li2023mts}.
Therefore, we envision a multi-model approach to pandemic forecasting in which specialized models on language,
temporal pattern and knowledge understating and processing cooperate with each other to produce accurate MTS pandemic forecasting.


\subsection{Knowledge Enhanced Graph Neural Networks}

One way to add knowledge to a model is to incorporate forecasting results from
science-based forecasting models.
For example, \cite{wang2020pm2} improved its Particulate Matter (${\rm PM}_{2.5}$)
GNN forecasting model by using weather forecasting results from the National Centers for Environmental Prediction (NCEP)'s Global Forecast System (GFS) and
climate forecasts from  the
European Centre for Medium-Range Weather Forecasts (ECMWF)'s ERA5.

The Knowledge Enhanced Graph Neural Network (KeGNN) \cite{werner2023knowledge} illustrates
another way to apply knowledge is by using a logic language.
It supports unary predicates for properties and binary predicates for relations.
The logic is mapped into real-valued vectors and functions.
A knowledge enhancement layer takes predictions (classification problem)
from a GNN and produces updated predictions based on how well the logic
is satisfied.

There is substantial research on using GNNs for Knowledge Graph Completion,
but little work in the opposite direction, creating GNNs from Knowledge Graphs
\cite{lin2020kgnn}.
For example, our research group has a project to utilize knowledge graphs
about COVID-19 to improve pandemic prediction.
Large Language Models can extract information from the scientific literature on COVID-19,
e.g., the COVID-19 Open Research Dataset Challenge (CORD-19).
\url{https://www.kaggle.com/datasets/allen-institute-for-ai/CORD-19-research-challenge}.
Knowledge Graph Embeddings (KGEs) can be used to transfer the
knowledge into the Forecasting Transformer.
Knowledge Embeddings may be concatenated with the input or
at some later stage of the transformer (referred to as knowledge infusion) \cite{hu2023survey}.
The self-attention mechanism of transformers can help to select the most useful knowledge.

\section{Meta Evaluation}

Multi-horizon forecasting is an important and challenging task, fraught with uncertainty.
Forecasts tend to degrade the longer the forecasting horizon.
The feasible horizon (the distance in the future a model can see before it degrades
to sheer speculation) varies with the domain and on what is being forecast.
For example, forecasting the weather (e.g. high and low daily temperature)
for a particular city, ten years from now is nonsense,
while forecasting the monthly Global Mean Surface Temperature (GMST)
or Pacific/Atlantic Sea Surface Temperature (SST)
with a horizon of 120 months can be done with reasonable accuracy with climate models
\cite{gordon2023unpredictable}.
Many of the papers address the problem of
Long Sequence Time series Forecasting (LSTF).

\subsection{Forecast Quality Metrics}

There are several metrics that can be used to evaluate the quality of a model's
forecasts as shown in Table \ref{tab:metrics}.
Given vectors ${\bf y}$, the actual observed values, and ${\bf \hat{y}}$, the forecasted values,
the following preliminary definitions are needed:
(1) A measure of variability of a time series, ${\rm var ({\bf y})}$,
(2) Random Walk (RW@h1) simply guesses the previous/latest value as the forecast and performs
reasonably well for horizon one forecasting, and thus can be used as a standard to measure other models.

\begin{small}
\begin{table}[htbp]
\caption{Forecast Quality Metrics}
\setlength{\tabcolsep}{4pt}
\begin{center}
\begin{tabular}{|c|c|c|} \hline
{\bf Metric} & {\bf Meaning}           & {\bf Formula}                                  \\ \hline \hline 
MSE   & Mean Squared Error             & ${\rm mean} \, (({\bf y} - {\bf \hat{y}})^2)$  \\ \hline
$R^2$ & Coefficient of Determination   & $1 - {\rm MSE} \,/\, {\rm var \, ({\bf y})}$   \\ \hline
RMSE  & Root Mean Squared Error        & $\sqrt{\rm MSE}$                               \\ \hline
NRMSE & Normalized RMSE                & ${\rm RMSE} / {\rm mean} \, ({\bf y})$         \\ \hline
MAE   & Mean Absolute Error            & ${\rm mean} \, (|{\bf y} - {\bf \hat{y}}|)$    \\ \hline
MAPE  & Mean Absolute Percentage Error & $100 \, {\rm mean} \left( \dfrac{|{\bf y} - {\bf \hat{y}}|}{|{\bf y}|} \right)$ \\ \hline
sMAPE & symmetric MAPE                 & $200 \, {\rm mean} \left( \dfrac{|{\bf y} - {\bf \hat{y}}|}{|{\bf y}| + |{\bf \hat{y}}|} \right)$ \\ \hline
MASE  & Mean Absolute Scaled Error     & $\dfrac{{\rm MAE (model)}}{{\rm MAE (RW@h1)}}$ \\ \hline
\end{tabular}
\label{tab:metrics}
\end{center}
\end{table}
\end{small}

\noindent
Notes: In the MAPE formula the mean/sum of ratios of the absolute values is the same as
the mean/sum of absolute values of ratios.
MSE, RMSE, and MAE require knowledge of the domains and their units to be interpretable.
NRMSE will blow up to infinity when the mean is zero, in which case the alternative definition
of dividing by the range could be used (but it is strongly affected by outliers).
$R^2$ is thought to be less informative than it is for regression problems,
due to temporal dependencies and changing values for variability over different time intervals.
MAPE has a tendency to blow up to infinity due to an observed zero value.
MAPE and sMAPE vary with units, for example, changing from Celsius to Kelvins
will make the errors look smaller.
MASE \cite{hyndman2006another} is a scale/unit invariant metric where a value of 1 means the model
is par with RW@1, less than 1 means it is better, and greater than 1 means it is worse
(but a MASE = 2 for horizon 10 may be quite good).

\subsection{Testing the Quality of Models}

Based on the paradigm of train and test, the ideal of $k$-fold cross-validation is not available
for time series due to temporal dependencies.
One should train on a beginning portion of the data and test points past the training portion.
A na\"ive way to do this would be to divide the dataset/time series, say 60\%-40\%,
and train on the first 60\% time points (to obtain values for model parameters)
and use these parameters along with training-only data to make very long horizon forecasts,
all the way through the testing set.
The forecasts will degrade because of staleness of (1) data and (2) parameter values.
The first problem can be fixed, by having a forecast horizon $h$ less than the size of the testing set.
Once forecasting through a window from 1 to $h$ is complete, move the window
ahead one time unit and redo the forecasts, maintaining all the forecasts in a matrix
where the row gives the time (e.g., the date) and the column gives the horizon
(e.g., the number of units/days ahead).
The redo will add one data point from the testing set to the values available for forecasting,
so the available information is freshened.
This leaves the staleness of the parameters which can be addressed by retraining,
by establishing a retraining frequency, say for every $10^{th}$ window, to retrain the parameters.
The training set would typically drop its first value and add the first value
from the testing set. 
For simple models, the retraining frequency can be high
(e.g., up to every time the window rolls), but to reduce high
computational costs for complex models, the frequency could be set lower.
Also, complex models may use incremental training to avoid
training from scratch every time.
Such techniques that reduce the staleness of data and/or parameters are
referred to as {\it rolling validation}.

For Foundation Models, more options are available as they come to the
problem already pre-trained.
One option would be to maintain the same process for rolling validation and
simply replace training with fine-tuning.

\subsection{Summary of Quality Findings from the Literature}

A meta-evaluation summarizes and establishes a consensus from the results
from relevant published literature.
This subsection summarizes results from recent papers that
compare MTS deep learning and foundation models on the following datasets:
Electricity Transformer Temperatures (ETT),
Influenza-Like Illness (ILI), and
Electricity Demand, Weather, and Traffic.

\begin{enumerate}

\item
{\bf ETTh\textsubscript{1,2}}
contains reporting of six electrical power loads, such as High Useful Load (HUFL) and High Useless Load (HULL),
and oil temperature of electricity transformers.
The data are recorded hourly for two transformers (1 and 2) in two counties in China over two years.

\item
{\bf ETTm\textsubscript{1,2}}
also contains reporting of power loads and oil temperature of electrical transformers recorded every 15 minutes.
Similar to the ETTh\textsubscript{1,2}, ETTm\textsubscript{1} includes the data for the first transformer,
while ETTm\textsubscript{2} contains the data for the second transformer.

\item
{\bf ILI}
contains weekly reporting of patients with symptoms related to influenza.
Each data time point includes variables, such as the age group, the number of reporting providers, and weighted and unweighted ILI cases.
The data are maintained by the Centers for Disease Control and Prevention (CDC) of the United States.

\item
{\bf Electricity}
contains the hourly electricity consumption of 321 customers.
The columns of the electricity dataset represent the customers.
Unlike the other datasets with multiple variables recorded concurrently, the electricity dataset has only one variable:
the hourly consumption recorded hourly for each customer.
The data are sourced from Elergone Energia in Portugal.

\item
{\bf Weather}
contains weather-related data, such as moisture level, CO\textsubscript{2} level, and amount of rain, among many other variables.
The data are maintained by the Max Planck Institute and collected by a weather station installed on the top roof of
the Institute for Biogeochemistry in Jena, Germany.

\item
{\bf Traffic}
contains a collection of road occupancy rates of a highway in San Francisco, California.
Similar to the electricity dataset, the columns of the traffic dataset represent the 862 sensors installed on the road;
it does not include any other variables, such as weather conditions, which may be helpful.
The occupancy rates are maintained by the Caltrans Performance Measurement System (PeMS) and are collected every hour.

\end{enumerate}

\begin{figure}[htbp]
    \includegraphics[width=0.7\textwidth, height=2.3in]{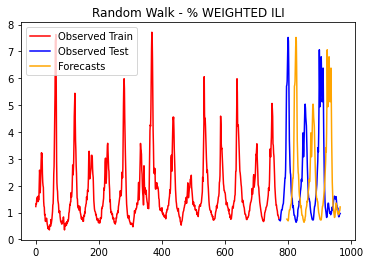}
    \caption{Percent of Weekly Patient Visits Exhibiting Influenza-Like Illness (ILI): Training (red), Testing (blue), RW (orange)}
    \label{fig:ili_data}
\end{figure}

To help ensure a fair comparison, every new proposed model uses the codebase made public by the Informer model \cite{zhou2021informer}
or one of the subsequent models based on it.
This often involves using the same data-loading, transformation, and evaluation experimental setup.
One, however, can observe some experimental or modeling differences, which can be summarized as follows:

\begin{itemize}

\item
Look-back window: Early works on utilizing the Transformer model for forecasting used a relatively short look-back window,
such as using a 96 look-back window to forecast 96, 192, 336, and 720 horizons for ETT datasets.
The authors of \cite{zeng2023transformers} explored increasing the look-back window for the Transformer-based models and found that
the forecasting performance tends to degrade, falling short in utilizing the extended long temporal information.
Alternatively, they showed that the PatchTST model benefits from the longer look-back window.

\item
Channel Mixing vs. Channel Independence:
In channel mixing, the attention is applied to all the features projected into a shared embedding space.
In channel-independence, the attention is applied separately to time points (or patches) in each channel, as in PatchTST, GPT4TS, and Time-LLM.
One argument in favor of channel-independence is that each channel behaves differently, which can harm the overall performance
when mixing all the channels.
In addition, the authors of \cite{zhang2022crossformer} explored using two-stage attention: one for temporal attention and
the other for cross-channel attention.

\item
Temporal Embeddings:
For both cases (channel mixing and channel-independence), a modeling technique projects the features into an embedding space,
and it can also use additional positional or temporal embeddings designed explicitly for each time frequency, such as weekly or hourly data.
For instance, the Informer model utilizes three types of embeddings: feature, positional, and temporal embeddings.
The PatchTST model uses two types: feature and positional.
The GPT4TS model utilizes only one: the feature embeddings.

\item
Instance Normalization:
Some modeling techniques also utilize the Reversible Instance Normalization (RevIN) \cite{kim2021reversible}, mitigating the
distribution shift between the look-back window and the forecasting horizons.
This technique simply normalizes the look-back window by subtracting the mean and dividing it by the standard deviation.
Subsequently, a modeling technique is trained on those normalized look-back windows to generate the forecasting horizons.
These forecasts then go through a denormalizing step to get the final forecasts for evaluation.
This simpler technique has been proven effective and contributes to much of the gain of the PatchTST model and GPT4TS.
For interested readers, please refer to Table 11 in the PatchTST paper for a detailed experiment on PatchTST with and without RevIN.

\item
Layer Normalization vs. Batch Normalization:
Most Transformer-based modeling techniques, such as the Informer or Time-LLM models, use Layer Normalization for the attention heads.
The PatchTST models, however, use batch normalization, which has been shown to improve the forecasting performance of the
time series transformers \cite{zerveas2021transformer}.

\item
Residual Attention:
In addition to the layer vs. batch normalization, the implementation of the attention block also varies across the Time Series Transformers.
For example, the PatchTST model uses residual attention, which keeps attention scores and adds them to the attention scores in the next layer.
The other models do not use any intermediate attention scores.

\item
Model Size:
The modeling techniques under study vary substantially in the model size (model dimensions and layers) used for modeling.
For example, the Informer uses two-layer attention blocks with the following parameters
(d\textsubscript{model}=512, n\textsubscript{heads}=8, and d\textsubscript{ff}=2048) for the ETTh1.
PatchTST uses a three layers attention block with these parameters
(d\textsubscript{model}=16, n\textsubscript{heads}=8, and d\textsubscript{ff}=128).
GPT4TS uses pre-trained GPT2 of the six layers with these parameters
(d\textsubscript{model}=768, n\textsubscript{heads}=4, and d\textsubscript{ff}=768) for the same dataset.

\item
On Baselines:
To some extent, the modeling techniques are limited in their evaluation since they do not include simpler baselines
that can sometimes be competitive, such as the Random Walk (RW), Mean Model (MM), or Simple Moving Average (SMA).
These simple baselines serve as good starting points as they do not require any training,
although SMA has a hyper-parameter (the window size or number of elements to average).
For example, Figure \ref{fig:ili_data} shows the Weighted (by age) Percentage of Patients with an ILI (for 966 time points)
that has been divided into train (772) and test (194) sets.
The forecasts (orange) are from the simple Random Walk (RW) baseline model and it tends to degrade
with increasing forecasting horizons, unless it happened to match a cycle/period.
In this case, the metrics for horizon = 24 weeks are
{\tt [ N = 170, MSE = 10.046, MAE = 2.473, sMAPE = 89.490 ]}.
Note, these results are on the original scale as {\tt sMAPE} values are distorted
when data are standardized (or normalized). A note on reproducibility, since the baseline results are often collected in each new proposed model without actually rerunning the other models,
it has been observed that some results are not reproducible by the research community in the GitHub repository of the models under study
(although the results (e.g., MAE) tend to be close).
In addition to the reproducibility, some researchers also observed a source of potential unfairness in the evaluations,
as some test time steps are omitted during the data loading where some models use different batch sizes (e.g., skip the
last incomplete batch).
Although the effects are small, it would simply follow-on work if a standard could be followed.
Furthermore, our testing has shown that small improvements in these models can be made by further tuning
of the hyper-parameters (suggesting it might be useful to provide a standardized setting and improved setting).
\end{itemize}
Table \ref{tab:evaluation} presents a comparative analysis of the modeling techniques under study
on eight benchmark datasets.
These models include
LLM-based models (GPT4TS),
Transformer-based models (PatchTST/42, FEDformer, Autoformer, Stationary, ETSformer, Informer, and Reformer),
CNN-based model (TimesNet), and
MLP-based models (NLinear, DLinear, and LightTS).
The results are sourced from \cite{zeng2023transformers, zhou2023one} and are generally consistent across different papers
where each newly proposed model does not run the baselines but collects results from a previous paper.
We included the models for which the code and hyper-parameters are made public,
omitting PatchTST/64 for which the hyper-parameters for ILI are not available.
The assessment employs MSE and MAE for the normalized observed and forecasted values,
averaged over all features and forecasting horizons.
The assessment shows that PatchTST consistently outperforms the other modeling techniques,
such as GPT4TS which is based on a pre-trained LLM.
PatchTST is also highly competitive with NLinear achieving comparable scores.
Table \ref{tab:evaluation} presents a rank of these models using the average MAE over all datasets and forecasting horizons.

One can observe that forecasting performance tends to be less variable
on the Electricity and Traffic datasets for all the modeling techniques including the Informer and the Reformer
models.
However, the forecasting performance tends to be highly variable on the ETT and ILI datasets
especially with Informer and Reformer being the least favorable models.

\begin{tiny}
\begin{table}[htbp]
\caption{Comparison of models using different look-back windows for forecasting \{24, 36, 48, 60\} horizons for ILI and \{96, 192, 336, 720\} for the other datasets. The evaluation metrics used are Mean Squared Error (MSE) and Mean Absolute Error (MAE) applied to the normalized observed and forecasted values. The lower the scores the better the forecast performance. The best scores are highlighted in \colorbox{gray!40}{\bf bold}. }
\setlength{\tabcolsep}{1.0pt}
\renewcommand{\arraystretch}{1.2}
\begin{center}
\begin{tabular}{|c@{\hspace{0.5pt}}c|cc|cc|cc|cc|cc|cc|cc|cc|cc|cc|cc|cc|}
\hline
 \multicolumn{2}{|c|}{Rank} & \multicolumn{2}{|c|}{3}  & \multicolumn{2}{c|}{2} & \multicolumn{2}{c|}{5} & \multicolumn{2}{c|}{1} & \multicolumn{2}{c|}{4} & \multicolumn{2}{c|}{8} & \multicolumn{2}{|c|}{9} & \multicolumn{2}{c|}{6} & \multicolumn{2}{c|}{7} & \multicolumn{2}{c|}{10} & \multicolumn{2}{c|}{11} & \multicolumn{2}{c|}{12}
 \\
\hline
 \multicolumn{2}{|c|}{Methods} & \multicolumn{2}{|c|}{GPT2(6)}  & \multicolumn{2}{c|}{NLinear} & \multicolumn{2}{c|}{DLinear} & \multicolumn{2}{c|}{PatchTST/42} & \multicolumn{2}{c|}{TimesNet} & \multicolumn{2}{c|}{FEDformer} & \multicolumn{2}{|c|}{Autoformer} & \multicolumn{2}{c|}{Stationary} & \multicolumn{2}{c|}{ETSformer} & \multicolumn{2}{c|}{LightTS} & \multicolumn{2}{c|}{Informer} & \multicolumn{2}{c|}{Reformer}
 \\
\hline  \hline
 &Metrics&MSE & MAE &MSE & MAE & MSE & MAE & MSE & MAE & MSE & MAE & MSE & MAE &
 MSE & MAE & MSE & MAE & MSE & MAE & MSE & MAE & MSE & MAE & MSE & MAE\\
\hline
\multirow{5}{*}{\rotatebox[origin=c]{90}{\tiny Weather}}
& 96 & 0.162 & 0.212 & 0.182&0.232&0.176 & 0.237 & 0.152 & 0.199 & 0.172 & 0.220 & 0.217 & 0.296 & 0.266 & 0.336 & 0.173 & 0.223 & 0.197 & 0.281 & 0.182 & 0.242 & 0.300 & 0.384 & 0.689 & 0.596 \\

& 192 & 0.204 & 0.248 &0.225&0.269& 0.220 & 0.282 & 0.197 & 0.243 & 0.219 & 0.261 & 0.276 & 0.336 & 0.307 & 0.367 & 0.245 & 0.285 & 0.237 & 0.312 & 0.227 & 0.287 & 0.598 & 0.544 & 0.752 & 0.638 \\

& 336 & 0.254 & 0.286 & 0.271&0.301& 0.265 & 0.319 & 0.249 & 0.283 & 0.280 & 0.306 & 0.339 & 0.380 & 0.359 & 0.395 & 0.321 & 0.338 & 0.298 & 0.353 & 0.282 & 0.334 & 0.578 & 0.523 & 0.639 & 0.596 \\

& 720 & 0.326 & 0.337 & 0.338&0.348& 0.333 & 0.362 & 0.320 & 0.335 & 0.365 & 0.359 & 0.403 & 0.428 & 0.419 & 0.428 & 0.414 & 0.410 & 0.352 & 0.288 & 0.352 & 0.386 & 1.059 & 0.741 & 1.130 & 0.792 \\

& Avg & 0.237 & 0.270 & 0.254&0.287&0.248 & 0.300 & \colorbox{gray!40}{\bf 0.229} & \colorbox{gray!40}{\bf 0.265} & 0.259 & 0.287 & 0.309 & 0.360 & 0.338 & 0.382 & 0.288 & 0.314 & 0.271 & 0.334 & 0.261 & 0.312 & 0.634 & 0.548 & 0.803 & 0.656 \\ \hline

\multirow{5}{*}{\rotatebox[origin=c]{90}{\tiny ETTh1}}
& 96 & 0.376 & 0.397 &0.374&0.394& 0.375 & 0.399 & 0.375 & 0.399 & 0.384 & 0.402 & 0.376 & 0.419 & 0.449 & 0.459 & 0.513 & 0.491 & 0.494 & 0.479 & 0.424 & 0.432 & 0.865 & 0.713 & 0.837 & 0.728 \\

& 192 & 0.416 & 0.418 & 0.408&0.415& 0.405 & 0.416 & 0.414 & 0.421 & 0.436 & 0.429 & 0.420 & 0.448 & 0.500 & 0.482 & 0.534 & 0.504 & 0.538 & 0.504 & 0.475 & 0.462 & 1.008 & 0.792 & 0.923 & 0.766 \\

& 336 & 0.442 & 0.433 &  0.429& 0.427&0.439 & 0.443 & 0.431 & 0.436 & 0.491 & 0.469 & 0.459 & 0.465 & 0.521 & 0.496 & 0.588 & 0.535 & 0.574 & 0.521 & 0.518 & 0.488 & 1.107 & 0.809 & 1.097 & 0.835 \\

& 720 & 0.477 & 0.456 & 0.440&0.453& 0.472 & 0.490 & 0.449 & 0.466 & 0.521 & 0.500 & 0.506 & 0.507 & 0.514 & 0.512 & 0.643 & 0.616 & 0.562 & 0.535 & 0.547 & 0.533 & 1.181 & 0.865 & 1.257 & 0.889 \\

& Avg & 0.427 & 0.426 & \colorbox{gray!40}{\bf0.413}&\colorbox{gray!40}{\bf0.422}&0.422 & 0.437 & 0.417 & 0.430 & 0.458 & 0.450 & 0.440 & 0.460 & 0.496 & 0.487 & 0.570 & 0.537 & 0.542 & 0.510 & 0.491 & 0.479 & 1.040 & 0.795 & 1.029 & 0.805 \\ \hline

\multirow{5}{*}{\rotatebox[origin=c]{90}{\tiny ETTh2}}
& 96 & 0.285 & 0.342 & 0.277&0.338& 0.289 & 0.353 & 0.274 & 0.336 & 0.340 & 0.374 & 0.358 & 0.397 & 0.346 & 0.388 & 0.476 & 0.458 & 0.340 & 0.391 & 0.397 & 0.437 & 3.755 & 1.525 & 2.626 & 1.317 \\

& 192 & 0.354 & 0.389 & 0.344&0.381& 0.383 & 0.418 & 0.339 & 0.379 & 0.402 & 0.414 & 0.429 & 0.439 & 0.456 & 0.452 & 0.512 & 0.493 & 0.430 & 0.439 & 0.520 & 0.504 & 5.602 & 1.931 & 11.12 & 2.979 \\

& 336 & 0.373 & 0.407 &  0.357& 0.400& 0.448 & 0.465 & 0.331 & 0.380 & 0.452 & 0.452 & 0.496 & 0.487 & 0.482 & 0.486 & 0.552 & 0.551 & 0.485 & 0.479 & 0.626 & 0.559 & 4.721 & 1.835 & 9.323 & 2.769 \\

& 720 & 0.406 & 0.441 & 0.394& 0.436&  0.605 & 0.551 & 0.379 & 0.422 & 0.462 & 0.468 & 0.463 & 0.474 & 0.515 & 0.511 & 0.562 & 0.560 & 0.500 & 0.497 & 0.863 & 0.672 & 3.647 & 1.625 & 3.874 & 1.697 \\

& Avg & 0.354 & 0.394 &0.343    &0.389&  0.431 & 0.446 & \colorbox{gray!40}{\bf0.330} & \colorbox{gray!40}{\bf0.379} & 0.414 & 0.427 & 0.437 & 0.449 & 0.450 & 0.459 & 0.526 & 0.516 & 0.439 & 0.452 & 0.602 & 0.543 & 4.431 & 1.729 & 6.736 & 2.191 \\ \hline

\multirow{5}{*}{\rotatebox[origin=c]{90}{\tiny ETTm1}}
& 96 & 0.292 & 0.346 & 0.306&0.348&  0.299 & 0.343 & 0.290 & 0.342 & 0.338 & 0.375 & 0.379 & 0.419 & 0.505 & 0.475 & 0.386 & 0.398 & 0.375 & 0.398 & 0.374 & 0.400 & 0.672 & 0.571 & 0.538 & 0.528 \\

& 192 & 0.332 & 0.372 &  0.349&0.375& 0.335 & 0.365 & 0.332 & 0.369 & 0.374 & 0.387 & 0.426 & 0.441 & 0.553 & 0.496 & 0.459 & 0.444 & 0.408 & 0.410 & 0.400 & 0.407 & 0.795 & 0.669 & 0.658 & 0.592 \\

& 336 & 0.366 & 0.394 & 0.375&0.388&  0.369 & 0.386 & 0.366 & 0.392 & 0.410 & 0.411 & 0.445 & 0.459 & 0.621 & 0.537 & 0.495 & 0.464 & 0.435 & 0.428 & 0.438 & 0.438 & 1.212 & 0.871 & 0.898 & 0.721 \\

& 720 & 0.417 & 0.421 & 0.433&0.422&  0.425 & 0.421 & 0.420 & 0.424 & 0.478 & 0.450 & 0.543 & 0.490 & 0.671 & 0.561 & 0.585 & 0.516 & 0.499 & 0.462 & 0.527 & 0.502 & 1.166 & 0.823 & 1.102 & 0.841 \\

& Avg & \colorbox{gray!40}{\bf0.352} & 0.383 &0.365&0.383& 0.388 & 0.403 & 0.357 & \colorbox{gray!40}{\bf0.378} & \colorbox{gray!40}{\bf0.352} & 0.382 & 0.400 & 0.406 & 0.448 & 0.452 &  0.481 & 0.456 & 0.429 & 0.425 & 0.435 & 0.437 & 0.961 & 0.734 & 0.799 & 0.671 \\ \hline

\multirow{5}{*}{\rotatebox[origin=c]{90}{\tiny ETTm2}}
& 96 & 0.173 & 0.262 & 0.167&0.255&  0.167 & 0.269 & 0.165 & 0.255 & 0.187 & 0.267 & 0.203 & 0.287 & 0.255 & 0.339 & 0.192 & 0.274 & 0.189 & 0.280 & 0.209 & 0.308 & 0.365 & 0.453 & 0.658 & 0.619 \\

& 192 & 0.229 & 0.301 & 0.221& 0.293&  0.224 & 0.303 & 0.220 & 0.292 & 0.249 & 0.309 & 0.269 & 0.328 & 0.281 & 0.340 & 0.280 & 0.339 & 0.253 & 0.319 & 0.311 & 0.382 & 0.533 & 0.563 & 1.078 & 0.827 \\

& 336 & 0.286 & 0.341 & 0.274& 0.327&  0.281 & 0.342 & 0.278 & 0.329 & 0.321 & 0.351 & 0.325 & 0.366 & 0.339 & 0.372 & 0.334 & 0.361 & 0.314 & 0.357 & 0.442 & 0.466 & 1.363 & 0.887 & 1.549 & 0.972 \\

& 720 & 0.378 & 0.401 & 0.368&0.384&  0.397 & 0.421 & 0.367 & 0.385 & 0.408 & 0.403 & 0.421 & 0.415 & 0.433 & 0.432 & 0.417 & 0.413 & 0.414 & 0.413 & 0.675 & 0.587 & 3.379 & 1.338 & 2.631 & 1.242 \\

& Avg & 0.266 & 0.326 &\colorbox{gray!40}{\bf0.257}&\colorbox{gray!40}{\bf0.315}& 0.267 & 0.333 & \colorbox{gray!40}{\bf0.257} & \colorbox{gray!40}{\bf0.315} & 0.291 & 0.333 & 0.305 & 0.349 & 0.327 & 0.371 & 0.306 & 0.347 & 0.293 & 0.342 & 0.409 & 0.436 & 1.410 & 0.810 & 1.479 & 0.915 \\ \hline

\multirow{5}{*}{\rotatebox[origin=c]{90}{\tiny ILI}}
& 24 & 2.063 & 0.881 & 1.683&0.858& 2.215 & 1.081 & 1.522 & 0.814 & 2.317 & 0.934 & 3.228 & 1.260 & 3.483 & 1.287 & 2.294 & 0.945 & 2.527 & 1.020 & 8.313 & 2.144 & 5.764 & 1.677 & 4.400 & 1.382 \\

& 36 & 1.868 & 0.892 & 1.703&0.859&  1.963 & 0.963 & 1.430 & 0.834 & 1.972 & 0.920 & 2.679 & 1.080 & 3.103 & 1.148 & 1.825 & 0.848 & 2.615 & 1.007 & 6.631 & 1.902 & 4.755 & 1.467 & 4.783 & 1.448 \\

& 48 & 1.790 & 0.884 & 1.719& 0.884& 2.130 & 1.024 & 1.673 & 0.854 & 2.238 & 0.940 & 2.622 & 1.078 & 2.669 & 1.085 & 2.010 & 0.900 & 2.359 & 0.972 & 7.299 & 1.982 & 4.763 & 1.469 & 4.832 & 1.465 \\

& 60 & 1.979 & 0.957 & 1.819& 0.917&  2.368 & 1.096 & 1.529 & 0.862 & 2.027 & 0.928 & 2.857 & 1.157 & 2.770 & 1.125 & 2.178 & 0.963 & 2.487 & 1.016 & 7.283 & 1.985 & 5.264 & 1.564 & 4.882 & 1.483 \\

& Avg & 1.925 & 0.903 &1.731&0.879&  2.169 & 1.041 & \colorbox{gray!40}{\bf1.538} & \colorbox{gray!40}{\bf0.841} & 2.139 & 0.931 & 2.847 & 1.144 & 3.006 & 1.161 & 2.077 & 0.914 & 2.497 & 1.004 & 7.382 & 2.003 & 5.137 & 1.544 & 4.724 & 1.445 \\ \hline

\multirow{5}{*}{\rotatebox[origin=c]{90}{\tiny Electricity}}
& 96 & 0.139 & 0.238 & 0.141& 0.237&  0.140 & 0.237 & 0.130 & 0.222 & 0.168 & 0.272 & 0.193 & 0.308 & 0.201 & 0.317 & 0.169 & 0.273 & 0.187 & 0.304 & 0.207 & 0.307 & 0.274 & 0.368 & 0.312 & 0.402 \\

& 192 & 0.153 & 0.251 & 0.154&0.248&  0.153 & 0.249 & 0.148 & 0.240 & 0.184 & 0.289 & 0.201 & 0.315 & 0.222 & 0.334 & 0.182 & 0.286 & 0.199 & 0.315 & 0.213 & 0.316 & 0.296 & 0.386 & 0.348 & 0.433 \\

& 336 & 0.169 & 0.266 & 0.171&0.265&  0.169 & 0.267 & 0.167 & 0.261 & 0.198 & 0.300 & 0.214 & 0.329 & 0.231 & 0.338 & 0.200 & 0.304 & 0.212 & 0.329 & 0.230 & 0.333 & 0.300 & 0.394 & 0.350 & 0.433 \\

& 720 & 0.206 & 0.297 & 0.210&0.297& 0.203 & 0.301 & 0.202 & 0.291 & 0.220 & 0.320 & 0.246 & 0.355 & 0.254 & 0.361 & 0.222 & 0.321 & 0.233 & 0.345 & 0.265 & 0.360 & 0.373 & 0.439 & 0.340 & 0.420 \\

& Avg & 0.167 & 0.263 &0.169&0.268& 0.166 & 0.263 & \colorbox{gray!40}{\bf0.162} & \colorbox{gray!40}{\bf0.253} & 0.192 & 0.295 & 0.214 & 0.327 & 0.227 & 0.338 & 0.193 & 0.296 & 0.208 & 0.323 & 0.229 & 0.329 & 0.311 & 0.397 & 0.338 & 0.422 \\ \hline

\multirow{5}{*}{\rotatebox[origin=c]{90}{\tiny Traffic}}
& 96 & 0.388 & 0.282 & 0.410&0.279&  0.410 & 0.282 & 0.367 & 0.251 & 0.593 & 0.321 & 0.587 & 0.366 & 0.613 & 0.388 & 0.612 & 0.338 & 0.607 & 0.392 & 0.615 & 0.391 & 0.719 & 0.391 & 0.732 & 0.423 \\

& 192 & 0.407 & 0.290 & 0.423&0.284&  0.423 & 0.287 & 0.385 & 0.259 & 0.617 & 0.336 & 0.604 & 0.373 & 0.616 & 0.382 & 0.613 & 0.340 & 0.621 & 0.399 & 0.601 & 0.382 & 0.696 & 0.379 & 0.733 & 0.420 \\

& 336 & 0.412 & 0.294 & 0.435&0.290&  0.436 & 0.296 & 0.398 & 0.265 & 0.629 & 0.336 & 0.621 & 0.383 & 0.622 & 0.337 & 0.618 & 0.328 & 0.622 & 0.396 & 0.613 & 0.386 & 0.777 & 0.420 & 0.742 & 0.420 \\

& 720 & 0.450 & 0.312 & 0.464&0.307&  0.466 & 0.315 & 0.434 & 0.287 & 0.640 & 0.350 & 0.626 & 0.382 & 0.660 & 0.408 & 0.653 & 0.355 & 0.632 & 0.396 & 0.658 & 0.407 & 0.864 & 0.472 & 0.755 & 0.423 \\

& Avg & 0.414 & 0.294 &0.433&0.289&  0.433 & 0.295 & \colorbox{gray!40}{\bf0.396} & \colorbox{gray!40}{\bf0.265} & 0.619 & 0.336 & 0.610 & 0.376 & 0.628 & 0.379 & 0.624 & 0.340 & 0.621 & 0.396 & 0.622 & 0.392 & 0.764 & 0.416 & 0.741 & 0.422 \\ \hline

& Avg All & 0.516 & 0.407 &0.490&0.400& 0.562 & 0.436 & \colorbox{gray!40}{\bf0.460} & \colorbox{gray!40}{\bf0.391} & 0.596 & 0.433 & 0.701 & 0.489 & 0.757 & 0.511 & 0.633 & 0.465 & 0.662 & 0.473 & 1.303 & 0.616 & 1.836 & 0.871 & 2.081 & 0.954  \\ \hline
\end{tabular}
\label{tab:evaluation}
\end{center}
\end{table}
\end{tiny}

\section{Summary}

The second wave of progress on multivariate time series forecasting has been recent and
appears to be heating up with the use of advanced deep learning architectures.
Work has started to establish further improvements using Knowledge Graphs
and Large Language Models.
Efforts have started to train them with scientific literature, such as the CORD-19
dataset for COVID-19.
A complete investigation of how this can help with time series forecasting
or related problems of time series classification or anomaly detection will take some time.
Analysis of the features/factors influencing the course/time evolution of a pandemic
may be conducted with the help of LLMs.
Findings from the analysis of a dataset (e.g., the effectiveness of mask policies)
can be quickly checked against previous studies.
New research can more quickly be suggested to fill gaps in existing knowledge.
Established knowledge can be associated with knowledge graphs or temporal knowledge graphs.
As discussed, there are multiple ways in which knowledge can be used
to improve forecasting.

Foundation models are being created for several modalities of data
and time series data are no exception.
During the Fall of 2023, several foundation models for time series data were
created and tested.
They show promise for providing more accurate and robust forecasts
as well as potential for greater explainability.
As discussed, there are multiple options for backbone models
as well as many choices for architectural elements.
Although successful elements from LLMs are a good starting point,
additional research is needed to optimize them for time series data.

Further potential is provided by multi-modal foundation models, where we emphasize the importance of modeling time series data with textual data.
It is challenging to build time series foundation models that are comparable to existing large language models since:
(1) unlike natural languages, existing time series data lacks the inherent semantic richness;
(2) the semantics in time series data are often heavily domain-specific
(e.g., the modeling of electrocardiogram signals would provide little help in predicting stock prices).
There are several potential benefits of the multi-model modeling:
(1) textual data can provide essential context that is not captured in the raw time series data;
(2) textual semantics can provide domain-specific knowledge that enhances the model’s interpretability;
(3) textual data can introduce additional features and dimensions of variability, which can help in training more robust and generalizable models.
In this way, text-enhanced time series models can be more easily transferred across different tasks
(e.g., classification, forecasting, anomaly detection) and domains.
Recently, foundation models like Contrastive Language-Image Pre-Training (CLIP) \cite{radford2021learning}
can effectively learn visual concepts from natural language supervision,
which shows the success of data augmentation with textual semantics.

The {\tt scalation/data} GitHub repository \url{https://github.com/scalation/data}
contains links to datasets, knowledge graphs, models, code, and papers on time series forecasting.
As the MSE and MAE values in this paper are based on normalized values and therefore hard to interpret
(other than by looking at the relative scores), 
tables in {\tt scalation/data} show MSE, MAE, and sMAPE quality metrics for the same forecasting models
on the original scale.
Results for COVID-19 pandemic prediction are also given in this repository.

\bibliographystyle{ACM-Reference-Format}
\bibliography{survey_paper}

\end{document}